\documentclass[]{article}
\usepackage{lmodern}
\usepackage{amssymb,amsmath}
\usepackage{ifxetex,ifluatex}
\usepackage{fixltx2e} 
\ifnum 0\ifxetex 1\fi\ifluatex 1\fi=0 
  \usepackage[T1]{fontenc}
  \usepackage[utf8]{inputenc}
\else 
  \ifxetex
    \usepackage{mathspec}
  \else
    \usepackage{fontspec}
  \fi
  \defaultfontfeatures{Ligatures=TeX,Scale=MatchLowercase}
\fi
\IfFileExists{upquote.sty}{\usepackage{upquote}}{}
\IfFileExists{microtype.sty}{%
\usepackage{microtype}
\UseMicrotypeSet[protrusion]{basicmath} 
}{}
\usepackage[margin=1in]{geometry}
\usepackage{hyperref}
\hypersetup{unicode=true,
            pdftitle={Empowering individual trait prediction using interactions},
            pdfauthor={Damian Gola Institut für Medizinische Biometrie und Statistik Universität zu Lübeck Universitätsklinikum Schleswig-Holstein, Campus Lübeck Lübeck, Germany gola@imbs.uni-luebeck.de; Inke R. König* Institut für Medizinische Biometrie und Statistik Universität zu Lübeck Universitätsklinikum Schleswig-Holstein, Campus Lübeck Lübeck, Germany inke.koenig@imbs.uni-luebeck.de},
            pdfborder={0 0 0},
            breaklinks=true}
\usepackage{longtable,booktabs}
\usepackage{graphicx,grffile}
\makeatletter
\def\maxwidth{\ifdim\Gin@nat@width>\linewidth\linewidth\else\Gin@nat@width\fi}
\def\maxheight{\ifdim\Gin@nat@height>\textheight\textheight\else\Gin@nat@height\fi}
\makeatother
\setkeys{Gin}{width=\maxwidth,height=\maxheight,keepaspectratio}
\IfFileExists{parskip.sty}{%
\usepackage{parskip}
}{
\setlength{\parindent}{0pt}
\setlength{\parskip}{6pt plus 2pt minus 1pt}
}
\providecommand{\tightlist}{%
  \setlength{\itemsep}{0pt}\setlength{\parskip}{0pt}}
\ifx\paragraph\undefined\else
\let\oldparagraph\paragraph
\renewcommand{\paragraph}[1]{\oldparagraph{#1}\mbox{}}
\fi
\ifx\subparagraph\undefined\else
\let\oldsubparagraph\subparagraph
\renewcommand{\subparagraph}[1]{\oldsubparagraph{#1}\mbox{}}
\fi

\let\rmarkdownfootnote\footnote%
\def\footnote{\protect\rmarkdownfootnote}

\usepackage{titling}

  \title{Empowering individual trait prediction using interactions}
    \pretitle{\vspace{\droptitle}\centering\huge}
  \posttitle{\par}
    \author{\textbf{Damian Gola}\\
Institut für Medizinische Biometrie und Statistik\\
Universität zu Lübeck\\
Universitätsklinikum Schleswig-Holstein, Campus Lübeck\\
Lübeck, Germany\\
\href{mailto:gola@imbs.uni-luebeck.de}{\nolinkurl{gola@imbs.uni-luebeck.de}} \\ \textbf{Inke R. König}*\\
Institut für Medizinische Biometrie und Statistik\\
Universität zu Lübeck\\
Universitätsklinikum Schleswig-Holstein, Campus Lübeck\\
Lübeck, Germany\\
\href{mailto:inke.koenig@imbs.uni-luebeck.de}{\nolinkurl{inke.koenig@imbs.uni-luebeck.de}}}
    \preauthor{\centering\large\emph}
  \postauthor{\par}
      \predate{\centering\large\emph}
  \postdate{\par}
    \date{* Corresponding author}

\usepackage{booktabs}
\usepackage{longtable}
\usepackage{array}
\usepackage{multirow}
\usepackage[table]{xcolor}
\usepackage{wrapfig}
\usepackage{float}
\usepackage{colortbl}
\usepackage{pdflscape}
\usepackage{tabu}
\usepackage{threeparttable}
\usepackage{threeparttablex}
\usepackage[normalem]{ulem}
\usepackage{makecell}

\usepackage{multirow}
\usepackage{booktabs}
\usepackage{threeparttable}
\usepackage{caption}
\usepackage{dsfont}
\usepackage{amsfonts}
\usepackage{placeins}

\begin{document}
\maketitle
\begin{abstract}
\textbf{Abstract}\\
\emph{Background} One component of precision medicine is to construct prediction models with their predictive ability as high as possible, e.g.~to enable individual risk prediction.
In genetic epidemiology, complex diseases like coronary artery disease, rheumatoid arthritis, and type 2 diabetes, have a polygenic basis and a common assumption is that biological and genetic features affect the outcome under consideration via interactions.
In the case of omics data, the use of standard approaches such as generalized linear models may be suboptimal and machine learning methods are appealing to make individual predictions.
However, most of these algorithms focus mostly on main or marginal effects of the single features in a dataset.
On the other hand, the detection of interacting features is an active area of research in the realm of genetic epidemiology.
One big class of algorithms to detect interacting features is based on the multifactor dimensionality reduction (MDR).
Here, we further develop the model-based MDR (MB-MDR), a powerful extension of the original MDR algorithm, to enable interaction empowered individual prediction.\\
\emph{Results} Using a comprehensive simulation study we show that our new algorithm can use information hidden in interactions more efficiently than two other state-of-the-art algorithms, namely the Random Forest and Elastic Net, and clearly outperforms these if interactions are present.
The performance of these algorithms is comparable if no interactions are present.
Further, we show that our new algorithm is applicable to real data by comparing the performance of the three algorithms on a dataset of rheumatoid arthritis cases and healthy controls.
As our new algorithm is not only applicable to biological/genetic data but to all datasets with discrete features, it may have practical implications in other applications as well, and we made our method available as an \texttt{R} package.\\
\emph{Conclusions} The explicit use of interactions between features can improve the prediction performance and thus should be included in further attempts to move precision medicine forward.

\end{abstract}

\FloatBarrier

\hypertarget{background}{%
\section{Background}\label{background}}

The concept of precision medicine can be viewed as a continuous process of data preprocessing/data mining (track 1), construction of diagnostic/prognostic models (track 2) and prediction of treatment response/disease progression (track 3) {[}\protect\hyperlink{ref-Konig2017}{1}{]}.
Whereas track 1 focuses on the identification of important observed and latent variables, tracks 2 and 3 require models with highly accurate predictions about disease status, prognosis or progression of a disease of a single individual {[}\protect\hyperlink{ref-Redekop2013}{2}--\protect\hyperlink{ref-Lin2017}{6}{]}.
Explained with (generalized) linear models as an example, based on the estimation of inference on regression coefficients tracks 2 and 3 aim at constructing models with their predictive ability as high as possible, measured in terms of some performance, e.g.~the area under the receiver operating characteristic curve (AUC).
In genetic epidemiology, simple Mendelian diseases, such as cystic fibrosis and hereditary breast and ovarian cancer, allow for relatively straightforward predictions. However, more complex diseases like coronary artery disease, rheumatoid arthritis, and type 2 diabetes, involve complex molecular mechanisms and thus have a polygenic basis {[}\protect\hyperlink{ref-Jordan2018}{7}{]}.
It is a common assumption that these biological/genetic features also are acting via interactions either with each other {[}\protect\hyperlink{ref-Cordell2002}{8}--\protect\hyperlink{ref-Ritchie2018}{11}{]} or with environmental features {[}\protect\hyperlink{ref-Eichler2010}{12}{]}.
To make it even more complicated, features may affect the outcome under consideration only via interactions.
Thus, the interacting features do not have an effect on their own.
An example of such a constellation is the effect of a variant in the \emph{MDR1} gene together with exposure to pesticides on Parkinson's disease {[}\protect\hyperlink{ref-Zschiedrich2009}{13}{]}.

The use of standard approaches such as generalized linear models is suboptimal in these cases because of the algorithm instabilities when modeling many variables and their interactions or requirements of large sample sizes {[}\protect\hyperlink{ref-Aschard2016}{14}{]}.
Thus, regularized generalized linear models {[}\protect\hyperlink{ref-Hoerl1970}{15}, \protect\hyperlink{ref-Tibshirani1996}{16}{]} or machine learning methods, e.g.~Random Forest {[}\protect\hyperlink{ref-Breiman2001}{17}{]}, are appealing to make individual predictions based on many variables.
They differ in the details, but most of them share one important property: they focus mostly on main or marginal effects of the single features in a dataset. For example, in Random Forest, at each node, the single best feature and its best split point are selected {[}\protect\hyperlink{ref-Ishwaran2015}{18}{]}.
This may lead to ignoring features without any or only small main effects, although Wright et al.~{[}\protect\hyperlink{ref-Wright2016}{19}{]} have shown that using enough single trees can compensate this issue.
Likewise, regularized regression models are usually specified using main effect terms only, and interaction terms have to be included explicitly as new features {[}\protect\hyperlink{ref-Friedman2010}{20}{]}.
These common limitations may limit the prediction performance of models based on currently used algorithms if features have an effect on the outcome only via interactions.

On the other hand, there has been much research on the detection of interacting features in the realm of genetic epidemiology {[}\protect\hyperlink{ref-Koo2015}{21}{]}.
One big class of algorithms to detect interacting features are the multifactor dimensionality reduction (MDR)-based algorithms based on the original idea by Ritchie et al.~{[}\protect\hyperlink{ref-Ritchie2001}{22}{]}.
The basic idea of all MDR-based algorithms is to reduce the dimensionality of simultaneously analyzed features by pooling combinations of feature levels (\emph{cells}) in high risk (\(H\)) and low risk (\(L\)) groups, resulting in a single best combination (\emph{MDR model}) of \(d\) features.
The original MDR algorithm has several drawbacks and limitations, thus a large number of modifications and extensions were proposed in recent years.
A comprehensive review of the original MDR algorithm and its modifications and extensions is given by Gola et al.~{[}\protect\hyperlink{ref-Gola2015}{23}{]}.
However, these algorithms aim at identifying the best classification model based on single and interacting features to identify interacting features but do not allow for individual predictions.

In this work we show how to extend the model-based MDR (MB-MDR), a powerful MDR-based algorithm to detect interacting features first described by Calle et al.~{[}\protect\hyperlink{ref-Calle2008a}{24}{]}, to enable interaction empowered individual prediction.

\FloatBarrier

\hypertarget{methods}{%
\section{Methods}\label{methods}}

\hypertarget{statistical-interactions}{%
\subsection{Statistical interactions}\label{statistical-interactions}}

In statistics, two variables are said to be statistically independent if the influence of them on the target size can be modeled as a function with two separate parameters for the two variables.
If an additional parameter is needed to describe the effect of both variables together on the target variable, this is referred to as interaction between the two variables in statistics.
The presence of a statistical interaction depends on the type of model used for modeling.
In the following, we will explain this in the light of a classification task in genetic epidemiology, i.e.~the prediction of a dichotomous outcome \(Y\) based on genetic markers, here given as single nucleotide polymorphisms (SNPs) with two alleles \(A_{1}\) and \(A_{2}\).

\hypertarget{definition-via-logistic-regression}{%
\subsubsection{Definition via logistic regression}\label{definition-via-logistic-regression}}

Cordell {[}\protect\hyperlink{ref-Cordell2009}{9}{]} defines the interaction of two SNPs \(A\) and \(B\) with alleles \(A_{1}\), \(A_{2}\) and \(B_{1}\), \(B_{2}\) on the basis of a saturated linear model, given by
\begin{align*}
l\left(\mathbb{E}\left(Y\right)\right) = &\beta_{0} + \beta^{A}_{A_{1}A_{2}}\mathds{1}_{A_{1}A_{2}} + \beta^{A}_{A_{2}A_{2}}\mathds{1}_{A_{1}A_{2}} + \beta^{B}_{B_{1}B_{2}}\mathds{1}_{B_{1}B_{2}} + \beta^{B}_{B_{1}B_{2}}\mathds{1}_{B_{1}B_{2}} + \\
&\iota_{A_{1}A_{2}B_{1}B_{2}}\mathds{1}_{A_{1}A_{2}B_{1}B_{2}} +  \iota_{A_{1}A_{2}B_{2}B_{2}}\mathds{1}_{A_{1}A_{2}B_{2}B_{2}} + \\
&\iota_{A_{2}A_{2}B_{1}B_{2}}\mathds{1}_{A_{2}A_{2}B_{1}B_{2}} +  \iota_{A_{2}A_{2}B_{2}B_{2}}\mathds{1}_{A_{2}A_{2}B_{2}B_{2}}.
\end{align*}

Here, \(l\) is a link function and, for a dichotomous outcome \(Y\) which follows a Bernoulli distribution with parameter \(\pi\), \(l\) is the logarithm of the odds of \(\pi=\mathbb{P}\left(Y=1\right)\):
\[
l\left(\mathbb{E}\left(Y\right)\right) = \mathrm{logit}\left(\pi\right) = \ln\left(\frac{\mathbb{P}\left(Y=1\right)}{1-\mathbb{P}\left(Y=1\right)}\right),
\]
\(\beta^{A}_{A_{1}A_{2}}\), \(\beta^{A}_{A_{2}A_{2}}\), \(\beta^{B}_{B_{1}B_{2}}\), \(\beta^{B}_{B_{2}B_{2}}\) are the main effects , and \(\iota_{A_{1}A_{2}B_{1}B_{2}}\), \(\iota_{A_{1}A_{2}B_{2}B_{2}}\), \(\iota_{A_{2}A_{2}B_{1}B_{2}}\), \(\iota_{A_{2}A_{2}B_{2}B_{2}}\) are the interaction effects.
The SNPs \(A\) and \(B\) are said to be interacting if at least one of the interaction effects is not equal to zero, i.e.~there is a deviation from additivity of the main effects.

To simplify the model, one can assume a recessive, dominant or additive genetic model for the SNPs.
In this case, the regression model simplifies to
\[
\mathrm{logit}\left(\pi\right) = \beta_{0} + \beta^{A}x^{A} + \beta^{B}x^{B} + \iota x^{A}x^{B},
\]
where
\[
x^{A} = \begin{cases}
1 \quad A_{2}A_{2}, \\
0 \quad \text{otherwise}
\end{cases}
\]
for a recessive genetic model,
\[
x^{A} = \begin{cases}
0 \quad A_{1}A_{1}, \\
1 \quad \text{otherwise}
\end{cases}
\]
for a dominant genetic model and
\[
x^{A} = \begin{cases}
0 \quad A_{1}A_{1}, \\
1 \quad A_{1}A_{2}, \\
2 \quad A_{2}A_{2}
\end{cases}
\]
for an additive model for SNP \(A\).
The definitions for SNP \(B\) are equivalent.

\hypertarget{definition-via-penetrances}{%
\subsubsection{Definition via penetrances}\label{definition-via-penetrances}}

\begin{table}
  \centering
  \caption{\label{tab:pentable} Penetrance table.}
  {
    \begin{tabular}{lcccc}
    \toprule
    & & \multicolumn{3}{c}{\textbf{Genotype at SNP $B$}} \\
    \cmidrule(lr){3-5}
    & & $B_1B_1$ & $B_1B_2$ & $B_2B_2$ \\
    \midrule
    \multirow{3}{*}{\textbf{Genotype at SNP $A$}} & $A_1A_1$ & 0.62 & 0.62 & 0.82 \\
    & $A_1A_2$ & 0.62 & 0.62 & 0.82 \\
    & $A_2A_2$ & 0.73 & 0.73 & 0.88 \\
    \bottomrule
    \end{tabular}
  }
  \begin{tablenotes}
    \item \textit{Note:}\\Example of a penetrance table for a multilocus ${\mathcal L}=\left(L_{1}, L_{2}\right)$ with two diallelic SNPs $L_{1}=A$ and $L_{2}=B$.
  \end{tablenotes}
\end{table}

The definition of statistical interaction is not limited to the deviation from the additivity of the main effects in the logistic model.
Another definition is based on the penetrance, which corresponds to the modeling of the probability \(\pi=\mathbb{P}\left(Y=1\right)\) by a linear model.
The penetrance function is defined as
\[
f^{A}\left(g\right) = \mathbb{P}\left(Y=1\mid G=g\right),
\]
i.e.~the probability of \(Y=1\) given a specific genotype \(g\in\left\lbrace A_{1}A_{1}, A_{1}A_{2}, A_{2}A_{2}\right\rbrace = \left\lbrace 0, 1, 2\right\rbrace\) of SNP \(A\).
If multiple SNPs \(L_{j}\), \(j=1,\dots,d\), \(d\in\mathbb{N}^{+}\) as multilocus \({\mathcal L}=\left(L_{1},\dots,L_{d}\right)\) are considered, the definition of the penetrance function can easily be extended to
\[
f^{{\mathcal L}}_{\boldsymbol{g}} := f^{{\mathcal L}}\left(g_{1},\dots,g_{d}\right) = \mathbb{P}\left(Y=1\mid G_{1} = g_{1}, \dots, G_{d}=g_{d}\right).
\]
Here, \(g_{j}\) is one of three possible genotypes of SNP \(L_{j}\) and \(\boldsymbol{g}=g_{1},\dots,g_{d}\) is one specific combination of genotypes.
Similar to the definition in the regression framework, an interaction between loci in terms of penetrance may be expressed as deviation from the additivity in the penetrance changes per allele.
As an example, assume two SNPs \(A\) and \(B\) with alleles \(A_{1}\), \(A_{2}\) and \(B_{1}\), \(B_{2}\).
Further, assume that at SNP \(A\) the penetracne increases from \(f^{A}\left(0\right) = f^{A}\left(1\right) = 0.62\) by 0.11 to \(f^{A}\left(2\right) = 0.73\) for genotype \(A_{2}A_{2}\) compared to the genotypes \(A_{1}A_{1}\) and \(A_{1}A_{2}\).
At the same time, the penetrance is increased from \(f^{B}\left(0\right) = f^{B}\left(1\right) = 0.62\) for the genotypes \(B_{1}B_{1}\) and \(B_{1}B_{2}\) by 0.2 to \(f^{B}\left(2\right) = 0.82\) if the genotype \(B_{2}B_{2}\) is present at SNP B.
If both genotypes \(A_{2}A_{2}\) and \(B_{2}B_{2}\) are present at the same time, a penetrance of \(f^{A,B}_{2,2} = 0.62 + 0.11 + 0.2 = 0.93\) would be expected.
However, in this example \(f^{A,B}_{2,2} = 0.88\) (see Table \ref{tab:pentable}).
Thus, there is a deviation of \(-0.05\) from additivity and therefore the SNPs \(A\) and \(B\) could be considered as interacting.

\begin{table}
  \centering
  \caption{\label{tab:marginalpentable} Penetrance table without any marginal effects.}
  \begin{tabular}{p{3.4cm}ccccc}
    \toprule
    & & \multicolumn{3}{c}{\textbf{Genotype at SNP $B$}} & \multirow{2}{2.8cm}{\textbf{Marginal penetrance of $A$}} \\
    \cmidrule(lr){3-5}
    & & $B_1B_1$ (0.25) & $B_1B_2$ (0.5) & $B_2B_2$ (0.25) \\
    \midrule
    \multirow{3}{*}{\textbf{Genotype at SNP $A$}} & $A_1A_1$ (0.25) & 0 & 0 & 1 & 0.25 \\
    & $A_1A_2$ (0.5) & 0 & 0.5 & 0 & 0.25 \\
    & $A_2A_2$ (0.25) & 1 & 0 & 0 & 0.25 \\
    \textbf{Marginal} & & \multirow{2}{*}{0.25} & \multirow{2}{*}{0.25} & \multirow{2}{*}{0.25} \\
    \textbf{penetrance of $B$} \\
    \bottomrule
  \end{tabular}
  \begin{tablenotes}
    \item \textit{Note:}\\Example of a penetrance table for a multilocus ${\mathcal L}=\left(L_{1}, L_{2}\right)$ with two diallelic SNPs $L_{1}=A$ and $L_{2}=B$ without any marginal effects.  
  \end{tablenotes}
\end{table}

If an SNP \(L_{j}\) falls outside the consideration of the multilocus \({\mathcal L}\), then the penetrances of the resulting multilocus genotypes are
\[
f_{g_{1},\dots,g_{j-1},g_{j+1},\dots,g_{d}}^{{\mathcal L}\setminus L_{j}} = \sum_{g_{j}}\mathbb{P}\left(G_{j}=g_{j}\right)f_{g_{1},\dots,g_{j-1},g_{j},g_{j+1},\dots,g_{d}}^{{\mathcal L}}.
\]
We refer to \(f_{g_{1},\dots,g_{j-1},g_{j+1},\dots,g_{d}}^{{\mathcal L}\setminus L_{j}}\) as marginal penetrances with respect to \(L_{j}\).
If
\[
f_{g_{1},\dots,g_{j-1},g_{j+1},\dots,g_{d}}^{{\mathcal L}\setminus L_{j}} = f^{{\mathcal L}\setminus L_{j}}\quad \forall g_{1},\dots,g_{j-1},g_{j+1},\dots,g_{d},
\]
then we say that \(L_{j}\) has no marginal effect.
An example of this concept is given in Table \ref{tab:marginalpentable} for a multilocus \({\mathcal L}\) with two diallelic SNPs.
The respective genotype probabilities are given in parentheses.
If SNP \(A\) is ignored, the penetrance on each genotype of SNP \(B\) is 0.25. Conversely, the penetrances of SNP \(A\) are also 0.25 if SNP \(B\) is ignored.
Thus, both SNPs have no marginal effect.
In this example, it is even the case that the marginal penetrances are identical with respect to both SNPs.
If these loci were analyzed independently, no effect would be identifiable.

Using penetrances two measures can be calculated, which give information about the influence of the considered multilocus \({\mathcal L}\) on an outcome \(Y\).
For this purpose Urbanowicz et al.~{[}\protect\hyperlink{ref-Urbanowicz2012b}{25}{]} calculate the population prevalence \(K\) as
\[
K = \sum_{\boldsymbol{g}\in\boldsymbol{G}}\mathbb{P}\left(\boldsymbol{g}\right)f^{{\mathcal L}}_{\boldsymbol{g}}.
\]
This expresses the probability with which a phenotype can be observed in a population.
Based on \(K\), the heritability
\[
h^{2} = \frac{1}{K\left(1-K\right)}\sum_{\boldsymbol{g}\in\boldsymbol{G}}\mathbb{P}\left(\boldsymbol{g}\right)\left(f_{\boldsymbol{g}}^{{\mathcal L}}-K\right)^{2}
\]
can also be calculated for \({\mathcal L}\) as a measure of the explained variance of \(Y\) by \({\mathcal L}\).
The larger \(h^{2}\in [0, 1]\), the better \(Y\) can be explained only by considering \({\mathcal L}\).

\hypertarget{connection-between-logistic-regression-and-penetrance}{%
\subsubsection{Connection between logistic regression and penetrance}\label{connection-between-logistic-regression-and-penetrance}}

\begin{table}
  \centering
  \caption{\label{tab:penoregressiontable} Log-odds of penetrances in Table \ref{tab:marginalpentable} for two diallelic SNPs $A$ and $B$.}
  {
    \begin{tabular}{lcccc}
    \toprule
    & & \multicolumn{3}{c}{\textbf{Genotype at SNP $B$}} \\
    \cmidrule(lr){3-5}
    & & $B_1B_1$ & $B_1B_2$ & $B_2B_2$ \\
    \midrule
    \multirow{3}{*}{\textbf{Genotype at SNP $A$}} & $A_1A_1$ & 0.5 & 0.5 & 1.5 \\
    & $A_1A_2$ & 0.5 & 0.5 & 1.5 \\
    & $A_2A_2$ & 1 & 1 & 2 \\
    \bottomrule
    \end{tabular}
  }
\end{table}

In order to support the description of the simulation setting, we give a short explanation on how to convert the effects on a logistic regression scale to penetrances and vice versa.

The logarithmic odds at the multilocus \({\mathcal L}\) can be calculated by
\begin{align*}
  \mathrm{logit}\left(f_{\boldsymbol{g}}^{{\mathcal L}}\right) = \beta_{0} + \sum_{k=1}^{i}\sum_{\boldsymbol{G}_{k}}\iota_{\boldsymbol{g}_{k}},
\end{align*}

given effect parameters \(\iota\). In this notation, \(\boldsymbol{G}_{k}\) stands for all \(k\) tuples of the genotype combinations of \(\boldsymbol{g}\), and \(0\leq i\leq d\) is the number of genotypes in \(\boldsymbol{g}\) different from the homozygous genotype with the most common allele at SNP \(L_{j}\), \(j = 1,\dots,d\).
In particular, for \(k = 1\), the \(\iota_{\boldsymbol{g}_{1}}\) are the corresponding main effects \(\beta_{g_{j}}\), \(g_{j}\in\boldsymbol{G}_{1}\).

With the inverse of the \(\mathrm{logit}\) function \(\mathrm{logit}^{-1}\left(z\right) = \mathrm{expit}\left(z\right) := \frac{\exp\left(z\right)}{1+\exp\left(z\right)}\) the penetrance for each genotype combination can be calculated from the effect parameters of the logistic model.

Conversely, from a given penetrance \(f_{\boldsymbol{g}}^{{\mathcal L}}\) for all genotype combinations \(\boldsymbol{g}\), the effect parameters \(\iota_{\boldsymbol{g}}\) of the corresponding logistic model can be calculated by
\begin{align*}
  \iota_{\boldsymbol{g}} &= \left(-1\right)^{j}\tilde{f}^{{\mathcal L}}_{\boldsymbol{0}} + \left(-1\right)^{j+1}\sum\tilde{f}^{{\mathcal L}}_{\boldsymbol{G}_{1}} + \dots + \left(-1\right)^{2j}\sum\tilde{f}^{{\mathcal L}}_{\boldsymbol{G}_{j}} \\
    &= \left(-1\right)^{j}\tilde{f}^{{\mathcal L}}_{\boldsymbol{0}} + \sum_{k=1}^{j}\left(\left(-1\right)^{j+k}\sum\tilde{f}^{{\mathcal L}}_{\boldsymbol{G}_{k}}\right).
\end{align*}

Here, \(\tilde{f}^{{\mathcal L}}_{\boldsymbol{g}} = \mathrm{logit}\left(f^{{\mathcal L}}_{\boldsymbol{g}}\right)\) and \(f_{\boldsymbol{0}}\) is the penetrance of the genotype combination of the homozygous genotypes with the most common allele at the respective SNP.

It should be noted at this point that an interaction effect at one level does not automatically imply an interaction effect at the other level.
This is called a scale effect {[}\protect\hyperlink{ref-Kraft2007}{26}{]} and discussed in detail by Gola et al.~{[}\protect\hyperlink{ref-Gola347096}{27}{]}.
As an example, Table \ref{tab:penoregressiontable} shows the combinations of penetrances transformed into effect parameters of the logistic regression model from Table \ref{tab:marginalpentable}.
It can be seen that there is no deviation from additivity at the level of logistic regression.
The logarithmic odds increase for the genotype \(A_{2}A_{2}\) at SNP A by \(\beta^{A} = 0.5\) and for the genotype \(B_{2}B_{2}\) at SNP B by \(\beta^{B}= 1\).
When combining both genotypes, the logarithmic odds increase by \(\beta^{A} + \beta^{B} + \iota = 0.5 + 1 + 0 = 1.5\).

This effect may also lead to the effect that a pure interaction of SNPs without any main or marginal effects of the single SNPs on one level being an impure interaction of SNPs with main or marginal effects on the other level and thus easier to detect by appropriate methods.

\hypertarget{model-based-multifactor-dimensionality-reduction-mb-mdr}{%
\subsection{Model-based multifactor dimensionality reduction (MB-MDR)}\label{model-based-multifactor-dimensionality-reduction-mb-mdr}}

The MB-MDR is an extension of the MDR such that the assignment of the cell labels, i.e.~the combinations of feature levels, is based on an appropriate statistical test and that each possible combination of features, i.e.~MDR model, is ranked by a test statistic.
Suppose each sample \(i\), \(i=1,\dots,n\) is characterized by \(q\) discrete features \(\boldsymbol{x}_{i}=\left(x_{1},\dots,x_{q}\right)\), with each feature \(x_{j}\), \(j=1,\dots,q\) having \(l_{j}\) levels.
The observed outcome of each sample is denoted by \(y_{i}\) and can be of arbitrary scale.
The core algorithm of the MB-MDR consists of five steps:

\begin{enumerate}
\def\labelenumi{\arabic{enumi}.}
\tightlist
\item
  Select \(d\leq q\) features \(x_{j_{k}}\) with \(l_{j_{k}}\), \(k=1,\dots,d\) levels (\(j_{k}\in\left\lbrace{1,\dots,q}\right\rbrace\)).
\item
  Arrange the samples based on the selected features in the \(d\)-dimensional space by grouping samples with the same level combinations of the \(d\) features into cells \(c_{m}\), \(m=1,\dots,\prod_{k}^{d}l_{j_{k}}\).
\item
  Perform appropriate hypothesis tests with test statistics \(T_{m}\) and \(p\) values \(p_{m}\), comparing the samples in each cell \(c_{m}\) with all other samples not in \(c_{m}\).
\item
  Assign a label to each cell \(c_{m}\) to construct the MDR model defined by the selected features based on the hypothesis test:
\end{enumerate}

\begin{itemize}
\tightlist
\item
  If less than \(n_{\text{min}}\) samples are in \(c_{m}\) or \(p_{m}\geq\alpha\), \(c_{m}\) has an ambiguous risk and is labeled as \(O\).
\item
  If at least \(n_{\text{min}}\) samples are in \(c_{m}\) and \(p_{m}<\alpha\), the value of \(T_{m}\) determines the label of \(c_{m}\) as high risk (\(H\)) or low risk (\(L\)).
\end{itemize}

\begin{enumerate}
\def\labelenumi{\arabic{enumi}.}
\setcounter{enumi}{4}
\tightlist
\item
  Derive a test statistic for the current MDR model by selecting the maximum test statistic of two appropriate hypothesis tests:
\end{enumerate}

\begin{itemize}
\tightlist
\item
  Test samples in high risk cells against all other samples.
\item
  Test samples in low risk cells against all other samples.
\end{itemize}

This core algorithm is repeated for all \(r=1,\dots,\binom{q}{d}\) possible combinations of \(d\) out of \(q\) features and possibly for several values of \(d\), constructing MDR models \(f_{d,r}\).
Finally, the MDR models can be sorted by their respective test statistic and using a permutation-based strategy, \(p\) values can be assigned to each MDR model.
Several improvements and extensions of this basic algorithm allow to analyze different outcomes, such as dichotomous {[}\protect\hyperlink{ref-Calle2008}{28}{]}, continuous {[}\protect\hyperlink{ref-MahachieJohn2011}{29}{]} and survival {[}\protect\hyperlink{ref-VanLishout2013}{30}{]} traits, or to adjust for covariates and lower order effects of the features of an MDR model {[}\protect\hyperlink{ref-MahachieJohn2012}{31}{]}.
A fast C++ implementation of the MB-MDR is available {[}\protect\hyperlink{ref-Lishout2015}{32}{]} and used in this work.

\hypertarget{extension-of-mb-mdr-to-individual-prediction}{%
\subsection{Extension of MB-MDR to individual prediction}\label{extension-of-mb-mdr-to-individual-prediction}}

We extended the MB-MDR algorithm to not only detect interactions between features but to allow individual predictions based on the MDR models.
It is important to note that each MDR model is a prediction model in itself using \(d\) features and that each cell of an MDR model includes the predicted outcome for the respective feature levels combination.
Thus, after the construction of the MDR models and selection of the \(s\) best MDR models, the prediction for a new sample is the aggregation of the characteristics of the cells the sample falls into.
In our framework, instead of calculating \(p\) values of the MDR models, \(s\) is determined by cross-validation during training.

Suppose a new sample \(i^{\ast}\) with features \(\boldsymbol{x}_{i^{\ast}}\).
Then, \(i^{\ast}\) is a member of one specific cell \(c_{m}\) in each of the best \(s\) MDR models \(f_{d,r}\), \(r=1,\dots,s\).
Different types of predictions are possible using different cell values and aggregations.

\begin{enumerate}
\def\labelenumi{\arabic{enumi}.}
\tightlist
\item
  \emph{Predicting a binary outcome, i.e, the classification task.}
\end{enumerate}

\begin{enumerate}
\def\labelenumi{\alph{enumi})}
\tightlist
\item
  \emph{Hard classification.}
  Count the number of MDR models in which \(i^{\ast}\) is a member of cells labeled as \(H\) and cells labeled as \(L\).
  Then, the estimated class of \(i^{\ast}\) is the most frequent cell label among the \(s\) best MDR models.
\item
  \emph{Probability estimation.}
  The natural estimate for the probability of being member of a specific class for a new sample \(i^{\ast}\), given the membership in a certain cell \(c_{m}\) of a MDR model \(f_{d,r}\), is the proportion of the specific class in that cell, regardless of whether if it is labeled as either \(H\) or \(L\).
  The simple average across the \(s\) MDR models with the highest test statistics results in an aggregated estimate of the probability of being a case.
  Here, \(O\) labeled cells may be treated in either of two ways:
\end{enumerate}

\begin{itemize}
\tightlist
\item
  \(O\) labeled cells are considered as missing values and thus are not considered in the aggregated estimate.
\item
  \(O\) labeled cells are included as the global estimate of the class probabilities in the training dataset.
\end{itemize}

\begin{enumerate}
\def\labelenumi{\arabic{enumi}.}
\setcounter{enumi}{1}
\tightlist
\item
  \emph{Predicting a continuous outcome, i.e., the regression task.}
  The same principle as in probability estimation applies to prediction in regression tasks for a continuous outcome.
  Here, the predicted outcome is given by the average of the mean outcome of training samples in the respective cells of the \(s\) highest ranked MDR models.
  Again, \(O\) labelled cells may be treated in either of two ways:
\end{enumerate}

\begin{itemize}
\tightlist
\item
  \(O\) labelled cells are considered as missing values and thus are not considered in the aggregated estimate.
\item
  \(O\) labelled cells are included as the global estimate of the mean outcome in the training dataset.
  Hard classification can be done by taking the most frequent cell label \(H\) or \(L\) among the \(s\) MDR models.
\end{itemize}

\begin{enumerate}
\def\labelenumi{\arabic{enumi}.}
\setcounter{enumi}{2}
\tightlist
\item
  \emph{General risk prediction.}
  Additionally, a score can be constructed by counting \(H\) cells as \(+1\), \(L\) cells as \(-1\) and \(O\) cells as 0.
  The higher the score of \(i^{\ast}\), the higher the risk of the specific outcome.
\end{enumerate}

The MB-MDR classification algorithm (MBMDRC) described so far has been implemented for classification tasks as function \texttt{MBMDRC} in the \texttt{R} package \texttt{MBMDRClassifieR} available on \texttt{GitHub} {[}\protect\hyperlink{ref-Gola2018}{33}{]}.

\hypertarget{simulation-study}{%
\subsection{Simulation study}\label{simulation-study}}

A simulation study was performed to compare our proposed algorithm with two state-of-the-art prediction algorithms, the Random Forest {[}\protect\hyperlink{ref-Breiman2001}{17}{]} and the Elastic Net {[}\protect\hyperlink{ref-Zou2005}{34}{]}, a generalization of the LASSO {[}\protect\hyperlink{ref-Tibshirani1996}{16}{]} and ridge regression {[}\protect\hyperlink{ref-Hoerl1970}{15}{]}, for classification tasks.
As implementations we utilized the \texttt{R} (version 3.3.1) {[}\protect\hyperlink{ref-RCoreTeam2015}{35}{]} packages \texttt{ranger} (version 0.8.1.300) {[}\protect\hyperlink{ref-Wright2017}{36}{]} and \texttt{glmnet} (version 2.0-5) {[}\protect\hyperlink{ref-Friedman2010}{20}{]}.
We considered eight scenarios with different numbers of SNPs or combinations of SNPs as effect feature components:

\begin{enumerate}
\def\labelenumi{\arabic{enumi}.}
\tightlist
\item
  one single SNP 
\item
  five single SNPs without interaction 
\item
  one interaction of two SNPs 
\item
  one interaction of two SNPs and three single SNPs without interactions 
\item
  two interactions of two SNPs each 
\item
  three interactions of two SNPs each 
\item
  three interactions of two SNPS each and three single SNPs without interactions 
\item
  one interaction of three SNPs and three single SNPs without interactions 
\end{enumerate}

The effect strength of each component was defined by the heritability \(h^{2}\in\left\lbrace{0.05,0.1,0.2}\right\rbrace\).
Additionally, SNPs without any effect were added such that each dataset contained \(q=100\) independently simulated SNPs in total.
The minor allele frequencies (MAF) of the effect SNPs was set to 0.1, 0.2 or 0.4.
The MAFs of the additional SNPs were randomly selected from \(\left(0.05,0.5\right)\).
All genotypes were simulated under the assumption of Hardy-Weinberg equilibrium.
The penetrance tables of interacting SNPs, i.e.~the probability of having a phenotype given a certain combination of genotypes, characterized by the heritability and MAFs of the SNPs were generated by the \texttt{GAMETES} software (version 2.1) {[}\protect\hyperlink{ref-Urbanowicz2012b}{25}{]} without any marginal effects of the interacting SNPs.
It was not possible to generate penetrance tables for \(h^{2}=0.2\) in scenario 8 with \texttt{GAMETES}, thus this setting is left out in the following.
The penetrance tables of single effect SNPs were created under the restriction of rendering \(\beta\) coefficients in a logistic regression model with an additive coding of the SNPs.
In scenarios with multiple SNP combinations, the single penetrances were aggregated on the \emph{logit} scale and transformed back to probabilities using the \emph{expit} transformation.
Phenotype, e.g.~disease status, of a sample, was determined by drawing from a Bernoulli distribution with the aggregated penetrance as phenotype probability.
We considered sample sizes of 200, 1000, 2000 and 10000 with equal numbers of cases and controls.
For each scenario and combination of simulation parameters 50 datasets \(D\) were created as replicates.

\begin{longtable}[]{@{}llll@{}}
\caption{\label{tab:hyperparameters} Hyperparameter spaces used for tuning.}\tabularnewline
\toprule
\begin{minipage}[b]{0.22\columnwidth}\raggedright
Algorithm\strut
\end{minipage} & \begin{minipage}[b]{0.22\columnwidth}\raggedright
Hyperparameter\strut
\end{minipage} & \begin{minipage}[b]{0.22\columnwidth}\raggedright
Description\strut
\end{minipage} & \begin{minipage}[b]{0.22\columnwidth}\raggedright
Values\strut
\end{minipage}\tabularnewline
\midrule
\endfirsthead
\toprule
\begin{minipage}[b]{0.22\columnwidth}\raggedright
Algorithm\strut
\end{minipage} & \begin{minipage}[b]{0.22\columnwidth}\raggedright
Hyperparameter\strut
\end{minipage} & \begin{minipage}[b]{0.22\columnwidth}\raggedright
Description\strut
\end{minipage} & \begin{minipage}[b]{0.22\columnwidth}\raggedright
Values\strut
\end{minipage}\tabularnewline
\midrule
\endhead
\begin{minipage}[t]{0.22\columnwidth}\raggedright
\texttt{glmnet}\strut
\end{minipage} & \begin{minipage}[t]{0.22\columnwidth}\raggedright
\texttt{alpha}\strut
\end{minipage} & \begin{minipage}[t]{0.22\columnwidth}\raggedright
Elastic net mixing parameter. \texttt{alpha=1} is the LASSO, \texttt{alpha=0} is the ridge penalty.\strut
\end{minipage} & \begin{minipage}[t]{0.22\columnwidth}\raggedright
\(\left\lbrace{0,0.25,0.5,0.75,1}\right\rbrace\)\strut
\end{minipage}\tabularnewline
\begin{minipage}[t]{0.22\columnwidth}\raggedright
\texttt{ranger}\strut
\end{minipage} & \begin{minipage}[t]{0.22\columnwidth}\raggedright
\texttt{num.trees}\strut
\end{minipage} & \begin{minipage}[t]{0.22\columnwidth}\raggedright
Number of trees\strut
\end{minipage} & \begin{minipage}[t]{0.22\columnwidth}\raggedright
1000\strut
\end{minipage}\tabularnewline
\begin{minipage}[t]{0.22\columnwidth}\raggedright
\strut
\end{minipage} & \begin{minipage}[t]{0.22\columnwidth}\raggedright
\texttt{mtry}\strut
\end{minipage} & \begin{minipage}[t]{0.22\columnwidth}\raggedright
Number of variables to possibly split at in each node.\strut
\end{minipage} & \begin{minipage}[t]{0.22\columnwidth}\raggedright
\(\left[1,100\right]\subset \mathbb{N}\)\strut
\end{minipage}\tabularnewline
\begin{minipage}[t]{0.22\columnwidth}\raggedright
\strut
\end{minipage} & \begin{minipage}[t]{0.22\columnwidth}\raggedright
\texttt{min.node.size}\strut
\end{minipage} & \begin{minipage}[t]{0.22\columnwidth}\raggedright
Minimal node size.\strut
\end{minipage} & \begin{minipage}[t]{0.22\columnwidth}\raggedright
\(\left[10,100\right]\subset\mathbb{N}\)\strut
\end{minipage}\tabularnewline
\begin{minipage}[t]{0.22\columnwidth}\raggedright
\texttt{MBMDRC}\strut
\end{minipage} & \begin{minipage}[t]{0.22\columnwidth}\raggedright
\texttt{min.cell.size}\strut
\end{minipage} & \begin{minipage}[t]{0.22\columnwidth}\raggedright
Minimum number of samples with a specific genotype combination to be statistically relevant. If less, a cell is automatically labelled as \(O\).\strut
\end{minipage} & \begin{minipage}[t]{0.22\columnwidth}\raggedright
\(\left[0,50\right]\subset\mathbb{N}\)\strut
\end{minipage}\tabularnewline
\begin{minipage}[t]{0.22\columnwidth}\raggedright
\strut
\end{minipage} & \begin{minipage}[t]{0.22\columnwidth}\raggedright
\texttt{alpha}\strut
\end{minipage} & \begin{minipage}[t]{0.22\columnwidth}\raggedright
Significance level used to determine \(H\), \(L\) and \(O\) label of a cell.\strut
\end{minipage} & \begin{minipage}[t]{0.22\columnwidth}\raggedright
\(\left(0.01,1\right)\subset\mathbb{R}\)\strut
\end{minipage}\tabularnewline
\begin{minipage}[t]{0.22\columnwidth}\raggedright
\strut
\end{minipage} & \begin{minipage}[t]{0.22\columnwidth}\raggedright
\texttt{adjustment}\strut
\end{minipage} & \begin{minipage}[t]{0.22\columnwidth}\raggedright
Adjustment for lower order marginal effects.\strut
\end{minipage} & \begin{minipage}[t]{0.22\columnwidth}\raggedright
\{\texttt{NONE}, \texttt{CODOMINANT}\}\strut
\end{minipage}\tabularnewline
\begin{minipage}[t]{0.22\columnwidth}\raggedright
\strut
\end{minipage} & \begin{minipage}[t]{0.22\columnwidth}\raggedright
\texttt{order}\strut
\end{minipage} & \begin{minipage}[t]{0.22\columnwidth}\raggedright
Number of SNPs to be considered in MDR models.\strut
\end{minipage} & \begin{minipage}[t]{0.22\columnwidth}\raggedright
\(\left\lbrace{1,2}\right\rbrace\)\strut
\end{minipage}\tabularnewline
\begin{minipage}[t]{0.22\columnwidth}\raggedright
\strut
\end{minipage} & \begin{minipage}[t]{0.22\columnwidth}\raggedright
\texttt{order.range}\strut
\end{minipage} & \begin{minipage}[t]{0.22\columnwidth}\raggedright
Use \texttt{order} as upper limit?\strut
\end{minipage} & \begin{minipage}[t]{0.22\columnwidth}\raggedright
\{\texttt{TRUE}, \texttt{FALSE}\}\strut
\end{minipage}\tabularnewline
\begin{minipage}[t]{0.22\columnwidth}\raggedright
\strut
\end{minipage} & \begin{minipage}[t]{0.22\columnwidth}\raggedright
\texttt{o.as.na}\strut
\end{minipage} & \begin{minipage}[t]{0.22\columnwidth}\raggedright
Use \(O\) labelled cells as \texttt{NA} or as the global probability/mean estimate.\strut
\end{minipage} & \begin{minipage}[t]{0.22\columnwidth}\raggedright
\{\texttt{TRUE}, \texttt{FALSE}\}\strut
\end{minipage}\tabularnewline
\bottomrule
\end{longtable}

For the benchmarking regarding the AUC of the three algorithms, we used the \texttt{mlr} framework (version 2.12) {[}\protect\hyperlink{ref-Bischl2016}{37}{]}.
Each dataset \(D\) was split into datasets \(D_{1}\) and \(D_{2}\) of the same size.
Tuning was performed with 5-fold cross-validation on \(D_{1}\) using the \texttt{R} package \texttt{mlrMBO} (version 1.1.0) {[}\protect\hyperlink{ref-Bischl2017}{38}{]} for 100 iterations with \texttt{ranger} (\texttt{ntrees}: 500, \texttt{mtry}: square root of the number of tuning hyperparameters) as the surrogate learner.
The hyperparameter spaces considered for tuning are shown in Table \ref{tab:hyperparameters} together with their respective descriptions.
After tuning, a prediction model with the tuned parameters was built on \(D_{1}\) and the prediction performance was calculated on \(D_{2}\) for each replicate.

\hypertarget{application-to-real-data}{%
\subsection{Application to real data}\label{application-to-real-data}}

We also compared the performance of the algorithms on the data by the North American Rheumatoid Arthritis Consortium (NARAC) dataset comprised of 1194 cases with rheumatoid arthritis and 868 controls, genotyped at 545,080 SNPs, which is described in detail by {[}\protect\hyperlink{ref-Amos2009}{39}{]}.
Previously, {[}\protect\hyperlink{ref-Liu2011}{41}{]} identified some putatively interacting loci in the HLA region on chromosome 6 in this dataset.
We removed SNPs and samples with high missing rates (\(>0.02\) and \(>0.1\) respectively) and selected all SNPs with MAF \(>0.1\) on chromosome 6 after LD pruning (window size: \(10^6\) SNPs, step size: 1 SNP, \(r^2\) threshold: 0.75).
As in the benchmarking on the simulated datasets, we used \texttt{mlr} and \texttt{mlrMBO} with the same settings as before in nested cross-validation with 10-fold outer cross-validation.

The underlying \texttt{R} code is available on request.

\FloatBarrier

\hypertarget{results}{%
\section{Results}\label{results}}

\hypertarget{simulation}{%
\subsection{Simulation}\label{simulation}}

\hypertarget{benchmark}{%
\subsubsection{Benchmark}\label{benchmark}}

\begin{figure}
\includegraphics[width=1\linewidth]{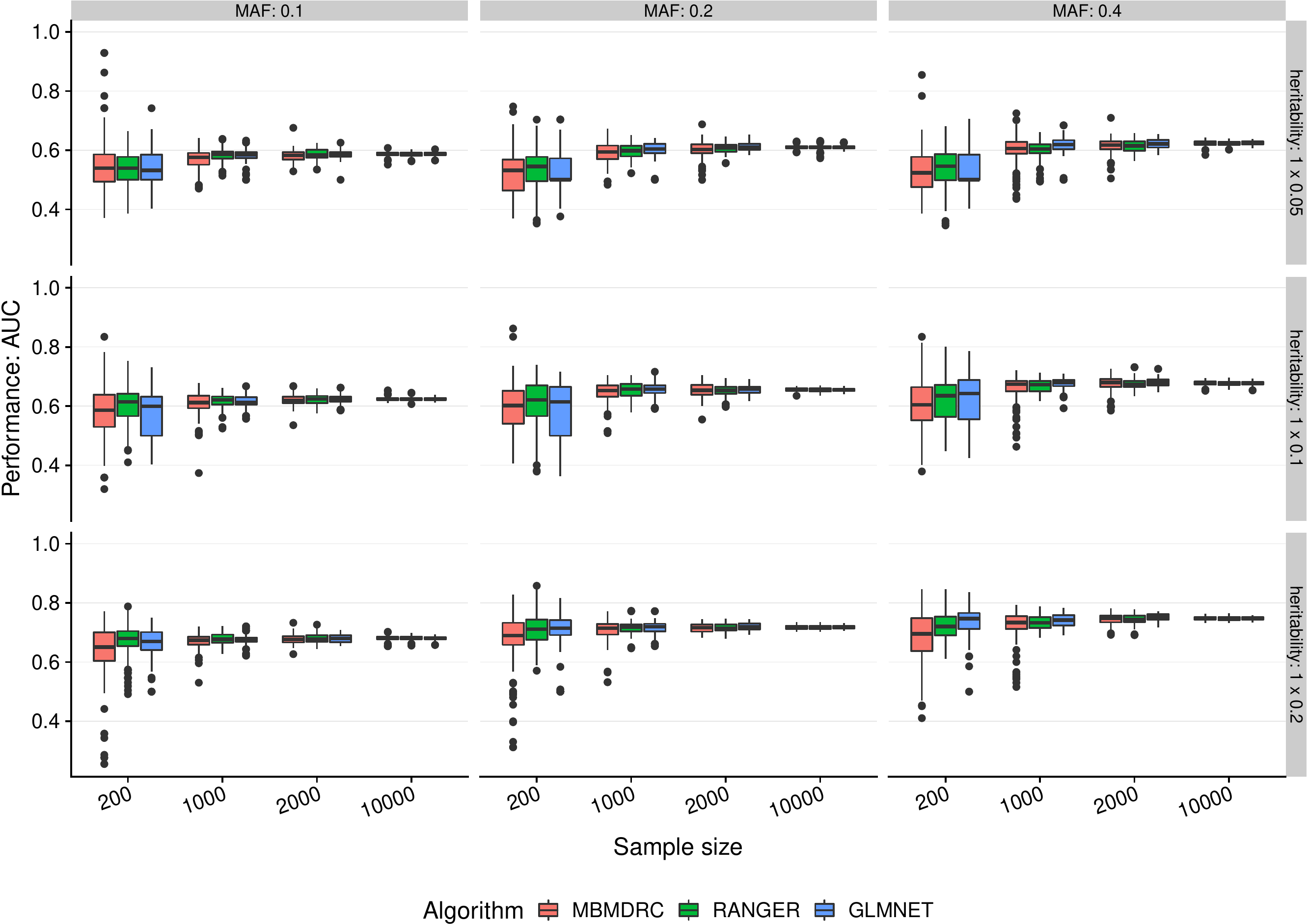} \caption{Performance in simulation scenario 1. Performance of the algorithms \texttt{MBMDRC}, \texttt{RANGER}, and \texttt{GLMNET} measured as AUC over 50 replicates in sample sizes 200, 1000, 2000, and 10000 in scenario 1 with one SNP with main effects (MAF 0.1, 0.2, or 0.4 and heritabilities 0.05, 0.1, 0.2.)}\label{fig:performance-plot-scenario1}
\end{figure}

\begin{figure}
\includegraphics[width=1\linewidth]{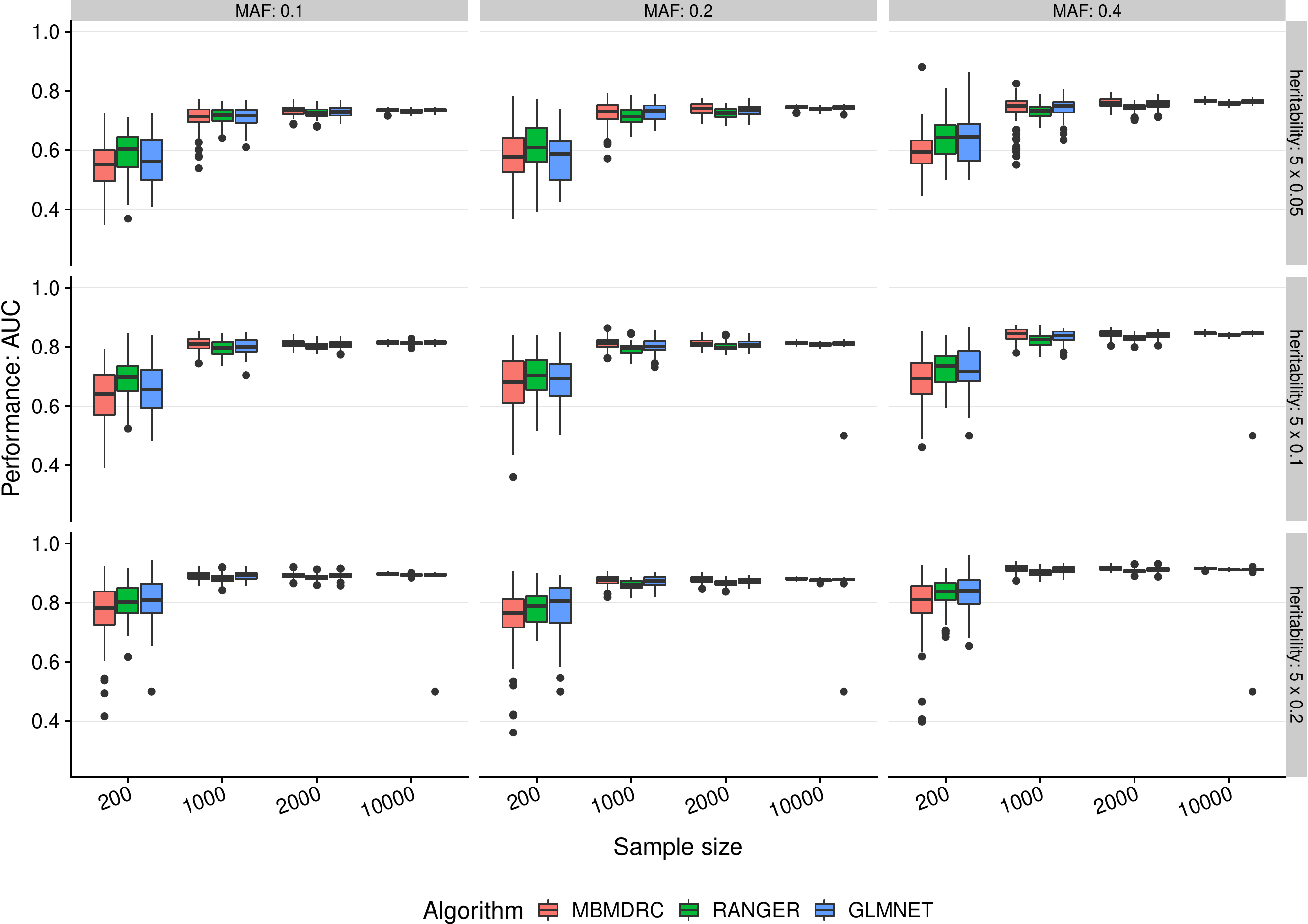} \caption{Performance in simulation scenario 2. Performance of the algorithms \texttt{MBMDRC}, \texttt{RANGER}, and \texttt{GLMNET} measured as AUC over 50 replicates in sample sizes 200, 1000, 2000, and 10000 in scenario 2 with five SNPs with main effects  (MAF 0.1, 0.2, or 0.4 and heritabilities 0.05, 0.1, 0.2).}\label{fig:performance-plot-scenario2}
\end{figure}

In scenarios 1 and 2 with only main effects simulated, all algorithms achieve similar performances (see Figures \ref{fig:performance-plot-scenario1} and \ref{fig:performance-plot-scenario2}).
All algorithms show the greatest variability of performance for the lowest sample sizes of 200, i.e.~100 for tuning and training and 100 for performance estimation.
With increasing sample size, the median performance increases and the variability of performance decreases for all algorithms.
The heritability has the greatest impact on prediction performance.
For example, the AUC increases from about 0.62 to 0.68 to 0.75 for heritabilities 0.05, 0.1 and 0.2 in scenario 1 and 10000 samples.
Comparing these two scenarios shows that the performance is dependent on the number of SNP combinations, i.e.~in these scenarios the number of SNPs with main effect.
This is expected, as the total heritability increases with the number of SNPs.

\begin{figure}
\includegraphics[width=1\linewidth]{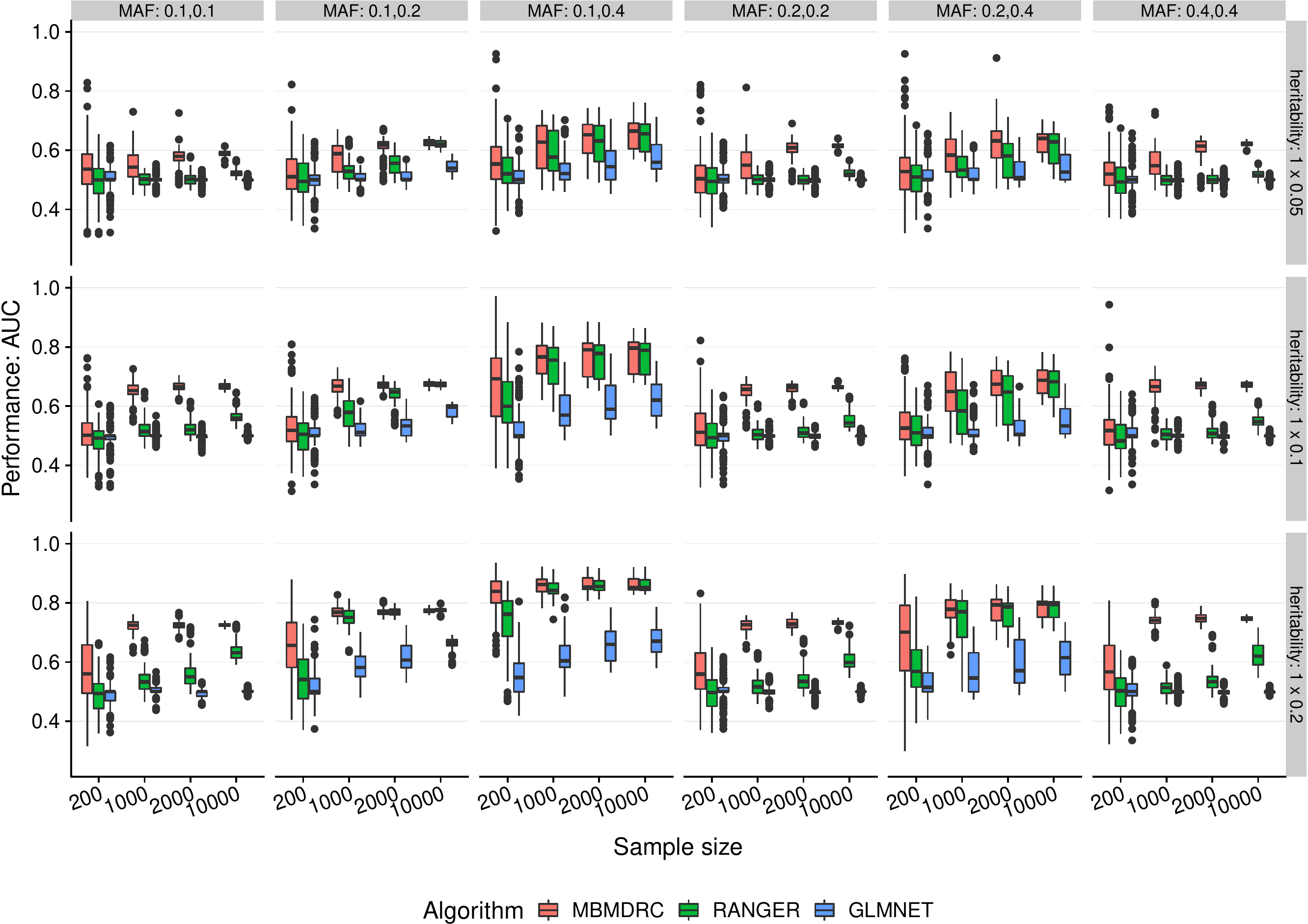} \caption{Performance in simulation scenario 3. Performance of the algorithms \texttt{MBMDRC}, \texttt{RANGER}, and \texttt{GLMNET} measured as AUC over 50 replicates in sample sizes 200, 1000, 2000, and 10000 in scenario 3 with one pair of interacting SNPs without marginal effects  (MAF 0.1, 0.2, or 0.4 and heritabilities 0.05, 0.1, 0.2.)}\label{fig:performance-plot-scenario3}
\end{figure}

Simulating only one interaction effect as in scenario 3, \texttt{MBMDRC} models have the highest median prediction performance for all sample sizes as shown in Figure \ref{fig:performance-plot-scenario3}.
In this scenario, the \texttt{ranger} and \texttt{glmnet} models do not achieve a median performance greater than 0.55, if the interacting SNPs have the same MAF.
Interestingly, models can improve their median performance at the cost of increased variability, if the SNPs have different MAFs.
This effect is most evident for \texttt{ranger} models, but also observable for the other two algorithms.

\begin{figure}
\includegraphics[width=1\linewidth]{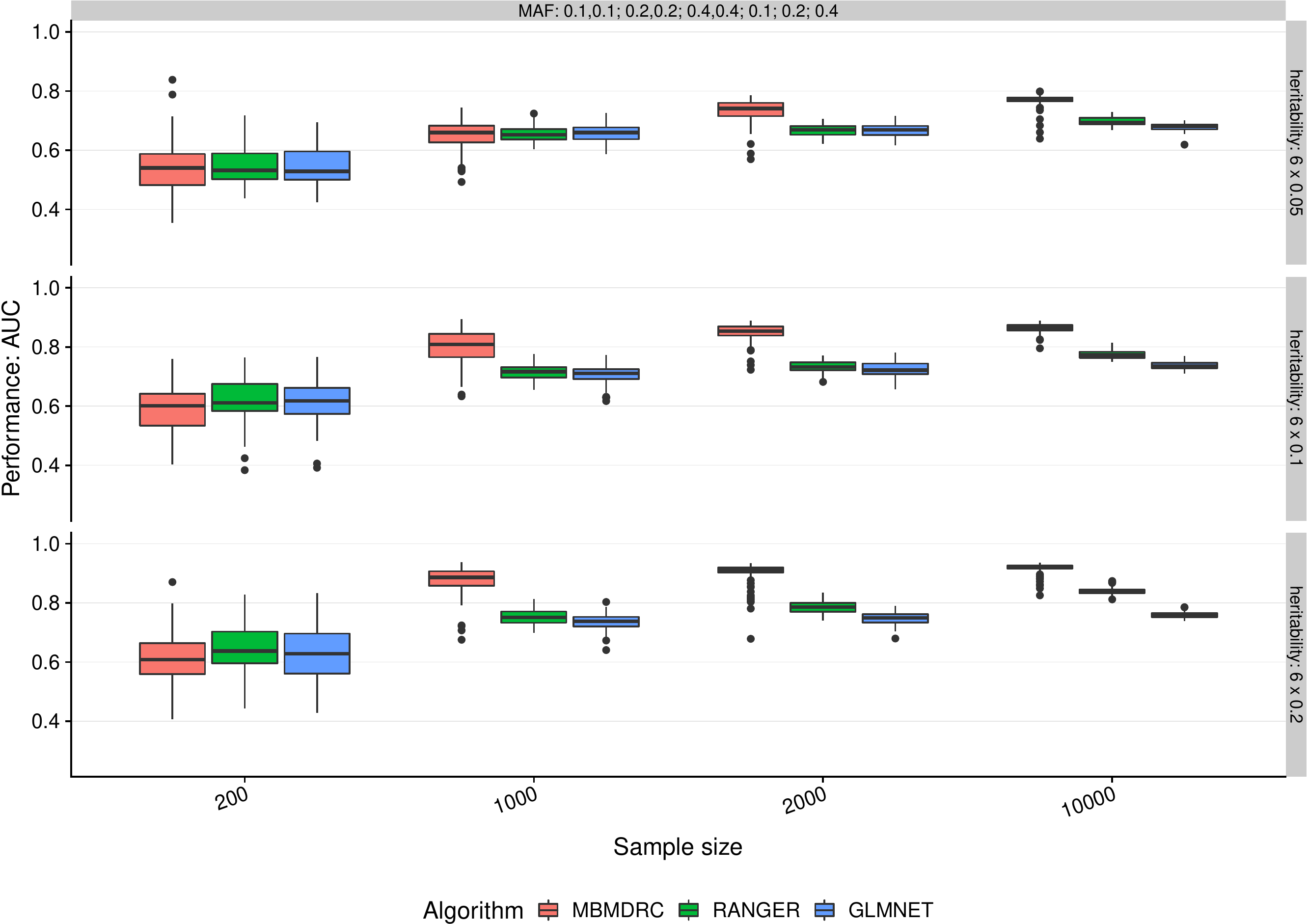} \caption{Performance in simulation scenario 7. Performance of the algorithms \texttt{MBMDRC}, \texttt{RANGER}, and \texttt{GLMNET} measured as AUC over 50 replicates in sample sizes 200, 1000, 2000, and 10000 in scenario 7 with three pairs of interacting SNPs without marginal effects and three SNPs with marginal effects only (MAF 0.1, 0.2, or 0.4 and heritabilities 0.05, 0.1, 0.2.)}\label{fig:performance-plot-scenario7}
\end{figure}

For scenarios with both main and interaction effects, \texttt{MBMDRC} models dominate the other two algorithms for sample sizes greater than 200 (see Figure \ref{fig:performance-plot-scenario7} for scenario 7).
However, \texttt{ranger} models can reach similar or even better performances if the interacting SNPs have different MAFs.
For example, for MAFs 0.1 and 0.4, heritability greater or equal 0.1 and sample size greater than 1000, \texttt{ranger} models achieve a better performance than the \texttt{MBMDRC} models in the median, although the variability is slightly increased.
The \texttt{glmnet} models do not use the interaction information, thus their performance is just based on the available main effects and the maximum median performances remain at about the same level between 0.68 and 0.80 as in scenario 1.

\begin{figure}
\includegraphics[width=1\linewidth]{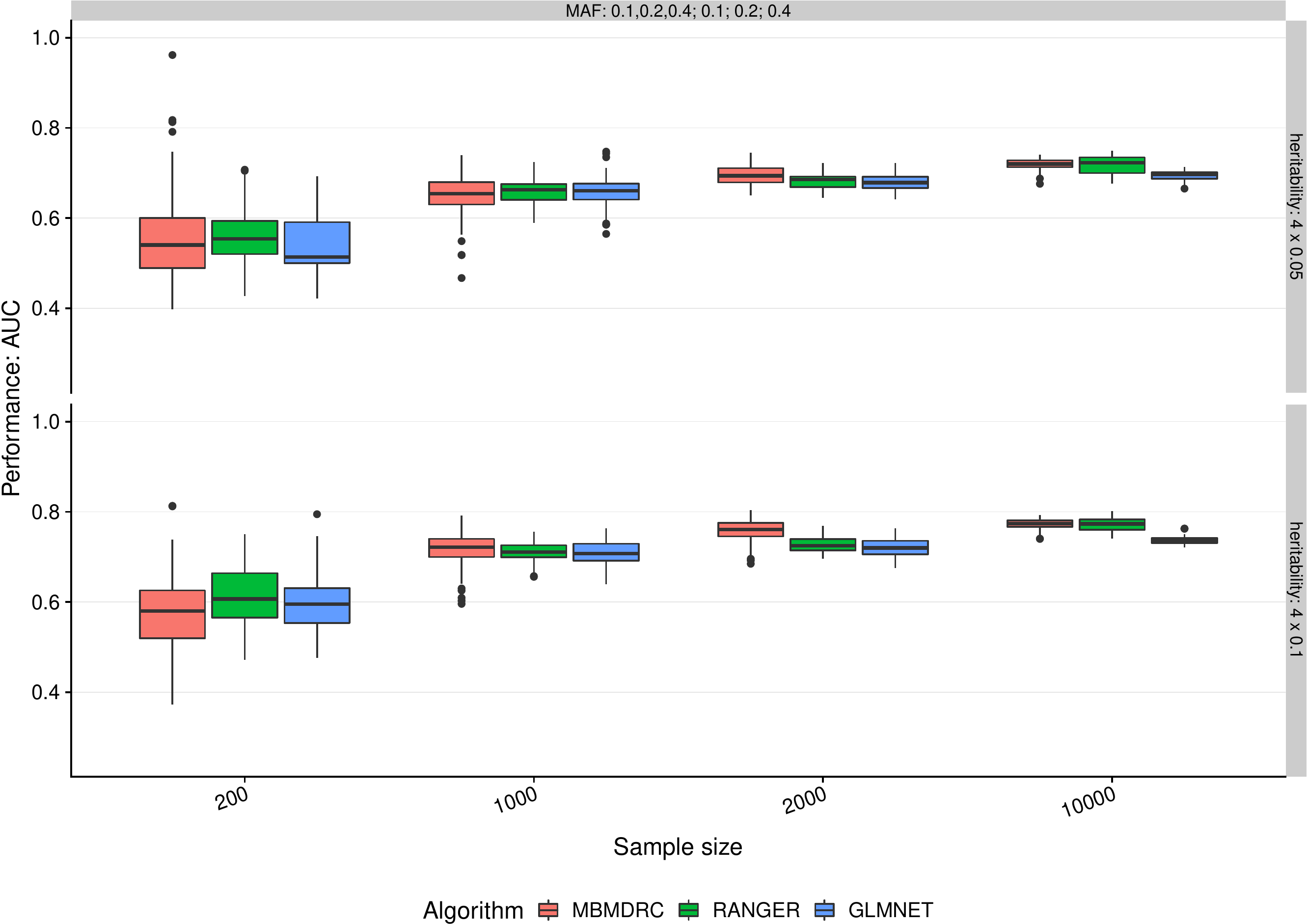} \caption{Performance in simulation scenario 8. Performance of the algorithms \texttt{MBMDRC}, \texttt{RANGER}, and \texttt{GLMNET} measured as AUC over 50 replicates in sample sizes 200, 1000, 2000, and 10000 in scenario 8 with three interacting SNPs without marginal effects and three SNPs with marginal effects only (MAF 0.1, 0.2, or 0.4 and heritabilities 0.05, 0.1, 0.2).}\label{fig:performance-plot-scenario8}
\end{figure}

In scenario 8, the interaction of three SNPs was simulated, wich is an interaction of higher order than considered by the MB-MDR.
Still, \texttt{MBMDRC} models achieve at least a similar performance as the \texttt{glmnet} and \texttt{ranger} models (see Figure \ref{fig:performance-plot-scenario8}).
Even though the \texttt{MBMDRC} models should be based mostly on the three additional single SNPs with marginal effects, the median performance for sample size 2000 is slightly better than those of \texttt{ranger} and \texttt{glmnet}.
For sample size 10000, \texttt{ranger} can achieve similar median performance but with higher variability.
The \texttt{glmnet} models are limited by the information based on the three main effects and their performance is comparable to those of scenario 7.

\begin{figure}
\includegraphics[width=1\linewidth]{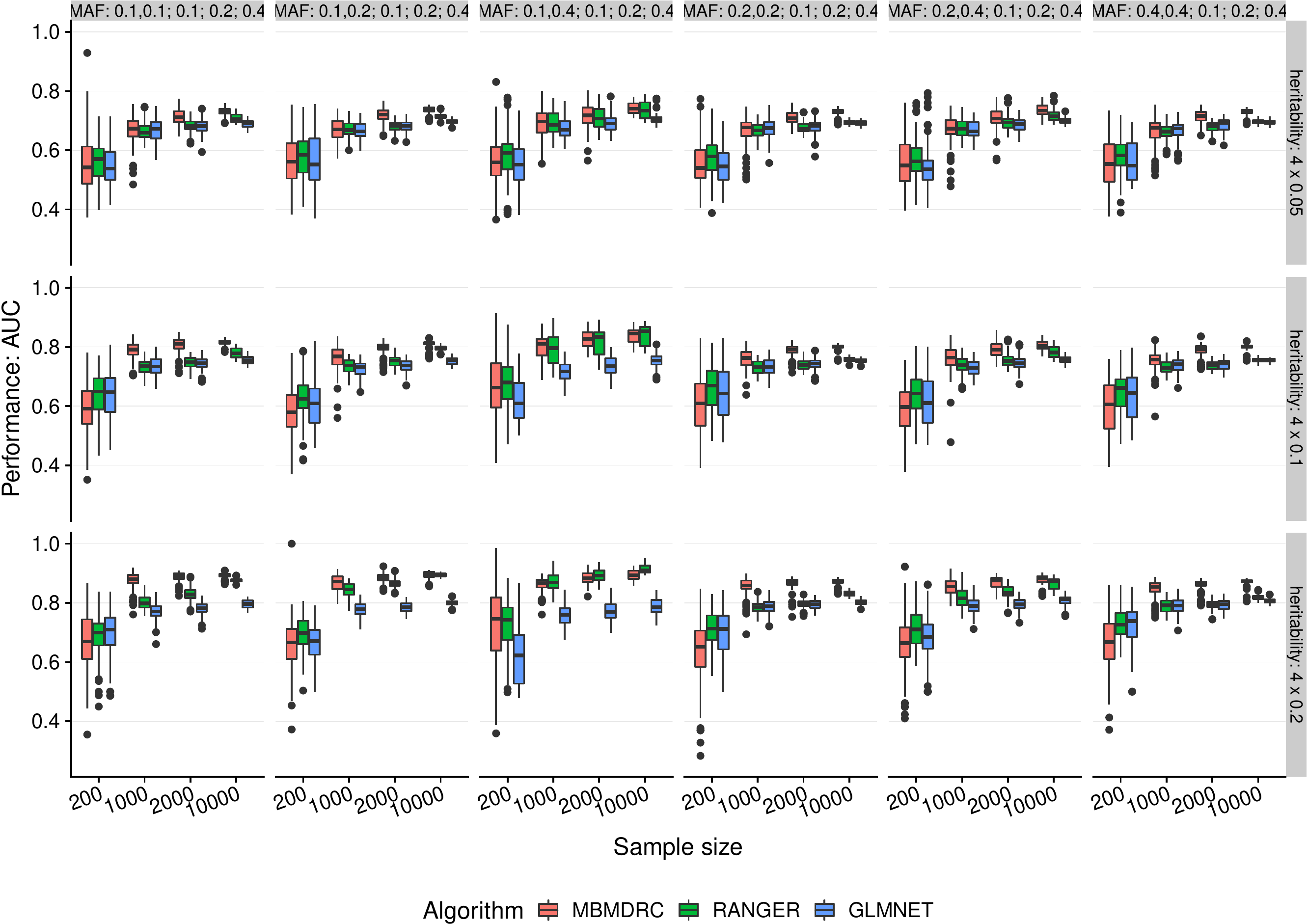} \caption{Performance in simulation scenario 4. Performance of the algorithms \texttt{MBMDRC}, \texttt{RANGER}, and \texttt{GLMNET} measured as AUC over 50 replicates in sample sizes 200, 1000, 2000, and 10000 in scenario 4 with one pair of interacting SNPs without marginal effects and three SNPs with main effects (MAF 0.1, 0.2, or 0.4 and heritabilities 0.05, 0.1, 0.2.)}\label{fig:performance-plot-scenario4}
\end{figure}

\begin{figure}
\includegraphics[width=1\linewidth]{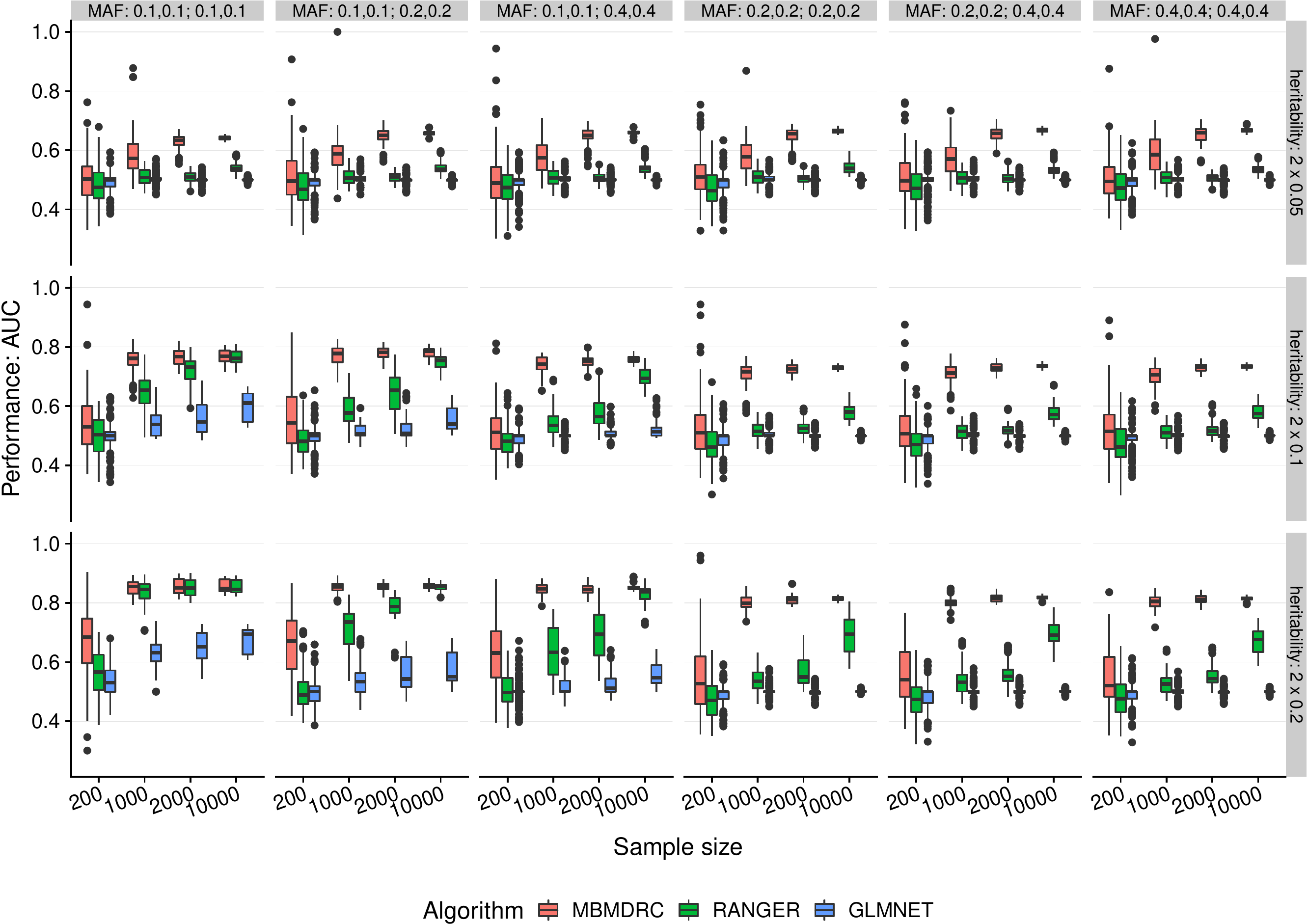} \caption{Performance in simulation scenario 5. Performance of the algorithms \texttt{MBMDRC}, \texttt{RANGER}, and \texttt{GLMNET} measured as AUC over 50 replicates in sample sizes 200, 1000, 2000, and 10000 in scenario 5 with two pairs of interacting SNPs without marginal effects  (MAF 0.1, 0.2, or 0.4 and heritabilities 0.05, 0.1, 0.2.)}\label{fig:performance-plot-scenario5}
\end{figure}

\begin{figure}
\includegraphics[width=1\linewidth]{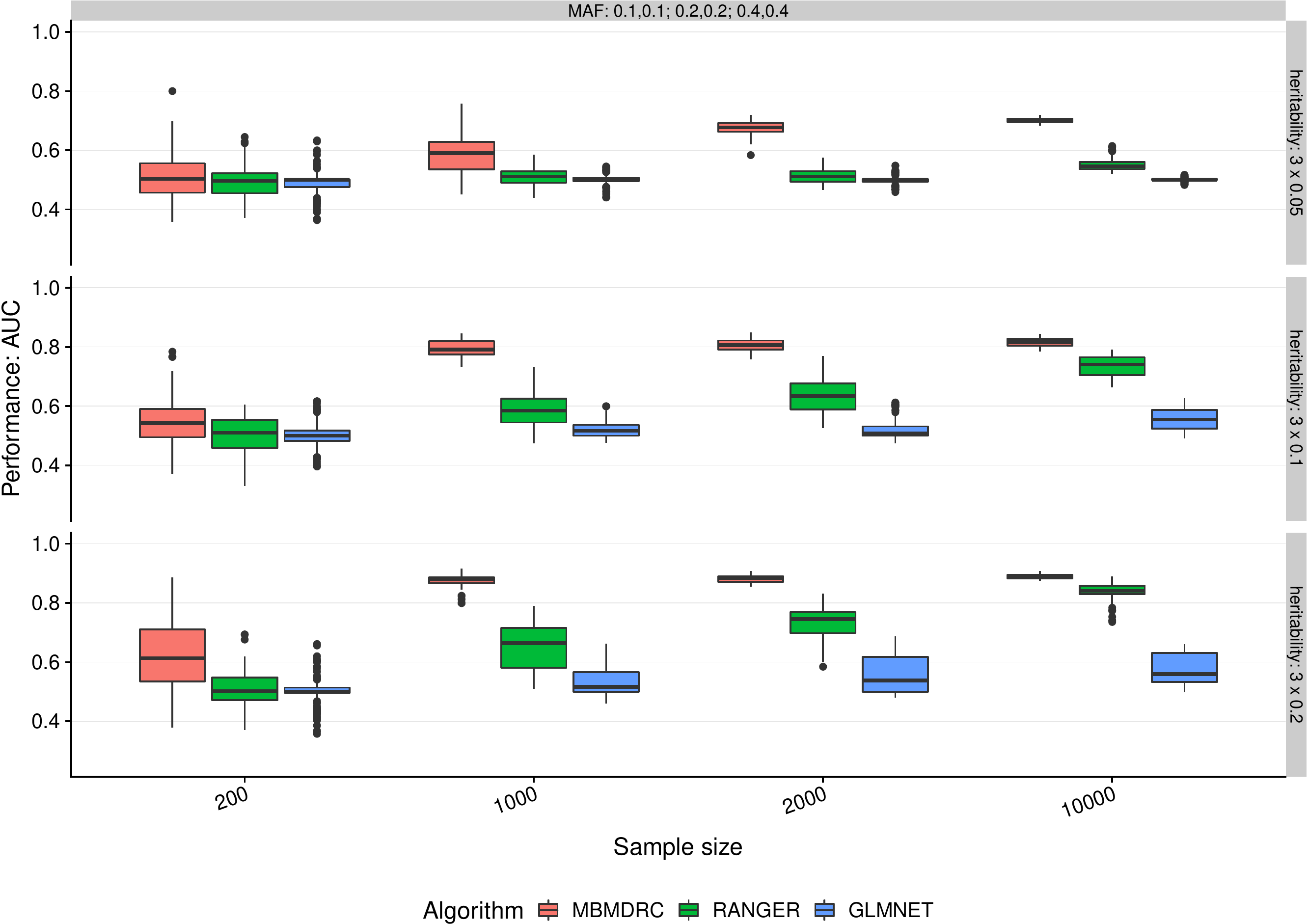} \caption{Performance in simulation scenario 6. Performance of the algorithms \texttt{MBMDRC}, \texttt{RANGER}, and \texttt{GLMNET} measured as AUC over 50 replicates in sample sizes 200, 1000, 2000, and 10000 in scenario 6 with three pairs of interacting SNPs without marginal effects  (MAF 0.1, 0.2, or 0.4 and heritabilities 0.05, 0.1, 0.2.)}\label{fig:performance-plot-scenario6}
\end{figure}

The further scenarios 4, 5, and 6 confirm the relationships described so far and the corresponding Figures \ref{fig:performance-plot-scenario4}, \ref{fig:performance-plot-scenario5}, and \ref{fig:performance-plot-scenario6} as well as detailed result tables also for the already described scenarios (Tables \ref{tab:performance-table-scenario1}, \ref{tab:performance-table-scenario2}, \ref{tab:performance-table-scenario3}, \ref{tab:performance-table-scenario4}, \ref{tab:performance-table-scenario5}, \ref{tab:performance-table-scenario6}, \ref{tab:performance-table-scenario7}, \ref{tab:performance-table-scenario8}) can be found in \protect\hyperlink{supplementary-information}{Supplementary Information}.

\hypertarget{selected-hyperparameters}{%
\subsubsection{Selected hyperparameters}\label{selected-hyperparameters}}

Inspection of the hyperparameters selected by tuning for the MBMDRC models shows that they correspond to the respective scenarios, i.e.~the scenario settings determine the optimal hyperparameter settings.
That, in turn, enables to select hyperparameter settings to include certain characteristics in prediction.

\begin{table}[t]

\caption{\label{tab:selected-hyperparameters-scenario1}Selected hyperparameters by tuning in scenario 1.}
\centering
\resizebox{\linewidth}{!}{
\begin{threeparttable}
\begin{tabular}{lrllll}
\toprule
\texttt{order\_range} & \texttt{order} & \texttt{adjustment} & Selection frequency (\%) & \texttt{alpha} & \texttt{min\_cell\_size}\\
\midrule
\texttt{FALSE} & 1 & \texttt{CODOMINANT} & 13.04 & 0.2744 (0.0950, 0.6319) & 26 (11, 38)\\
\texttt{FALSE} & 1 & \texttt{NONE} & 4.44 & 0.2631 (0.0771, 0.6532) & 25 (8, 42)\\
\texttt{FALSE} & 2 & \texttt{CODOMINANT} & 3.65 & 0.5040 (0.2370, 0.7539) & 23 (14, 43)\\
\texttt{FALSE} & 2 & \texttt{NONE} & 29.48 & 0.4472 (0.1883, 0.7452) & 18 (7, 34)\\
\texttt{TRUE} & 1 & \texttt{CODOMINANT} & 6.26 & 0.2199 (0.0590, 0.5655) & 23 (10, 39)\\
\addlinespace
\texttt{TRUE} & 1 & \texttt{NONE} & 3.39 & 0.2694 (0.0816, 0.6624) & 27 (8, 40)\\
\texttt{TRUE} & 2 & \texttt{CODOMINANT} & 17.83 & 0.4257 (0.1727, 0.7420) & 25 (13, 38)\\
\texttt{TRUE} & 2 & \texttt{NONE} & 21.91 & 0.4382 (0.1663, 0.7345) & 20 (8, 37)\\
\bottomrule
\end{tabular}
\begin{tablenotes}
\item \textit{Note: } 
\item For hyperparameters \texttt{alpha} and \texttt{min\_cell\_size} the median and the first and third quartile in parantheses are given.
\end{tablenotes}
\end{threeparttable}}
\end{table}

\begin{table}[t]

\caption{\label{tab:selected-hyperparameters-scenario3}Selected hyperparameters by tuning in scenario 3.}
\centering
\resizebox{\linewidth}{!}{
\begin{threeparttable}
\begin{tabular}{lrllll}
\toprule
\texttt{order\_range} & \texttt{order} & \texttt{adjustment} & Selection frequency (\%) & \texttt{alpha} & \texttt{min\_cell\_size}\\
\midrule
\texttt{FALSE} & 1 & \texttt{CODOMINANT} & 2.67 & 0.3112 (0.0986, 0.6948) & 29 (15, 45)\\
\texttt{FALSE} & 1 & \texttt{NONE} & 1.37 & 0.3781 (0.1633, 0.7290) & 28 (14, 46)\\
\texttt{FALSE} & 2 & \texttt{CODOMINANT} & 30.89 & 0.5152 (0.2113, 0.8135) & 13 (5, 29)\\
\texttt{FALSE} & 2 & \texttt{NONE} & 19.23 & 0.5267 (0.2331, 0.7976) & 12 (4, 28)\\
\texttt{TRUE} & 1 & \texttt{CODOMINANT} & 2.14 & 0.3411 (0.1078, 0.6796) & 29 (15, 45)\\
\addlinespace
\texttt{TRUE} & 1 & \texttt{NONE} & 1.34 & 0.2683 (0.0915, 0.5463) & 30 (13, 45)\\
\texttt{TRUE} & 2 & \texttt{CODOMINANT} & 24.17 & 0.5132 (0.2033, 0.7986) & 12 (4, 27)\\
\texttt{TRUE} & 2 & \texttt{NONE} & 18.19 & 0.5055 (0.2098, 0.7961) & 14 (5, 30)\\
\bottomrule
\end{tabular}
\begin{tablenotes}
\item \textit{Note: } 
\item For hyperparameters \texttt{alpha} and \texttt{min\_cell\_size} the median and the first and third quartile in parantheses are given.
\end{tablenotes}
\end{threeparttable}}
\end{table}

\begin{table}[t]

\caption{\label{tab:selected-hyperparameters-scenario7}Selected hyperparameters by tuning in scenario 7.}
\centering
\resizebox{\linewidth}{!}{
\begin{threeparttable}
\begin{tabular}{lrllll}
\toprule
\texttt{order\_range} & \texttt{order} & \texttt{adjustment} & Selection frequency (\%) & \texttt{alpha} & \texttt{min\_cell\_size}\\
\midrule
\texttt{FALSE} & 1 & \texttt{CODOMINANT} & 4.67 & 0.5347 (0.1374, 0.7596) & 20 (10, 36)\\
\texttt{FALSE} & 1 & \texttt{NONE} & 2.61 & 0.3341 (0.0806, 0.5735) & 14 (7, 30)\\
\texttt{FALSE} & 2 & \texttt{CODOMINANT} & 6.78 & 0.5620 (0.2932, 0.7883) & 18 (5, 38)\\
\texttt{FALSE} & 2 & \texttt{NONE} & 6.44 & 0.4080 (0.1837, 0.6850) & 14 (5, 23)\\
\texttt{TRUE} & 1 & \texttt{CODOMINANT} & 3.50 & 0.2639 (0.0942, 0.5165) & 15 (8, 29)\\
\addlinespace
\texttt{TRUE} & 1 & \texttt{NONE} & 2.17 & 0.3347 (0.1032, 0.6110) & 22 (12, 34)\\
\texttt{TRUE} & 2 & \texttt{CODOMINANT} & 67.83 & 0.3983 (0.1705, 0.6853) & 16 (7, 31)\\
\texttt{TRUE} & 2 & \texttt{NONE} & 6.00 & 0.5597 (0.2392, 0.8168) & 16 (5, 30)\\
\bottomrule
\end{tabular}
\begin{tablenotes}
\item \textit{Note: } 
\item For hyperparameters \texttt{alpha} and \texttt{min\_cell\_size} the median and the first and third quartile in parantheses are given.
\end{tablenotes}
\end{threeparttable}}
\end{table}

For scenarios 1, 3, and 7, Tables \ref{tab:selected-hyperparameters-scenario1}, \ref{tab:selected-hyperparameters-scenario3}, and \ref{tab:selected-hyperparameters-scenario7} summarize how often specific combinations of the hyperparameters \texttt{order\_range}, \texttt{order} und \texttt{adjustment} were selected in tuning.
These hyperparameters depend on each other, i.e.~interaction effects of two SNPs without any marginal effects can enter the prediction model only if \texttt{order} is set to \texttt{2}.
Simultaneously, the hyperparameters \texttt{order\_range} and \texttt{adjustment} influence whether additional marginal effects can be detected by the MB-MDR algorithm and thus used for prediction.
Since setting \texttt{adjustment} to \texttt{CODOMINANT} will adjust for possible marginal effects, \texttt{order\_range} should be set to \texttt{TRUE}, to allow for marginal effects.
On the other hand, if \texttt{order} is set to \texttt{1}, both \texttt{adjustment} and \texttt{order\_range} become meaningless.

In scenario 1, \texttt{order} is set to \texttt{2} in 72.87\% of the replicates, although no interactions are simulated in this scenario.
However, if the hyperparameters \texttt{order\_range} and \texttt{order} are set to \texttt{FALSE} and \texttt{2}, respectively, the hyperparameter \texttt{adjustment} is set to \texttt{NONE} more often (29.48\%) than to \texttt{CODOMINANT} (3.65\%).
Thus, the MB-MDR algorithms also accounts for marginal effects.
This is not the case, if \texttt{order\_range} and \texttt{order} are set to \texttt{TRUE} and \texttt{2}, respectively.
In this setting, \texttt{adjustment} is almost equally frequent set to \texttt{NONE} (21.91\%) and \texttt{CODOMINANT} (17.83\%).
The hyperparameter \texttt{alpha} seems to be dependent on the hyperparameter \texttt{order}.
If \texttt{order} is set to \texttt{1} the median of \texttt{alpha} between 0.22 and 0.27 whereas the median when \texttt{order} set to \texttt{2} ranges higher between 0.43 and 0.50.
However, the range of the quartiles is rather big, thus there seems to be no optimal value for \texttt{alpha} in general for this scenario.
Same is true for the hyperparameter \texttt{min\_cell\_size} with median between 18 and 27 and wide quartile ranges.

In scenario 3, \texttt{order} is set to \texttt{2} in 92.48\% of the replicates, as expected, since in this scenario, only interaction effects are simulated.
Although there are no marginal effects of single SNPs simulated in this scenarios, there is a slight tendency to set \texttt{adjustment} to \texttt{CODOMINANT}, whereas the selection frequencies of the values \texttt{TRUE} (45.84\%) and \texttt{FALSE} (54.16\%) for the hyperparameter \texttt{order\_range} are almost equal.
Again, the hyperparameters \texttt{alpha} and \texttt{min\_cell\_size} depend on \texttt{order} and no optimal values can be determined because of the wide range of the quartiles.

In scenario 7 with simulated marginal and interaction effects, one combination of hyperparameters stands out.
In this scenario the hyperparameters \texttt{order}, \texttt{order\_range}, and \texttt{adjustment} are set to \texttt{2}, \texttt{TRUE}, and \texttt{CODOMINANT} in 59.14\% of the replicates, much more frequent than any other combination of hyperparameter values.
Using these hyperparameter settings, marginal and interaction effects can clearly be differentiated and used for prediction.
Again, no clear optimal values for the hyperparameters \texttt{alpha} and \texttt{min\_cell\_size} can be determined.

\hypertarget{real-data}{%
\subsection{Real data}\label{real-data}}

\begin{figure}
\includegraphics[width=1\linewidth]{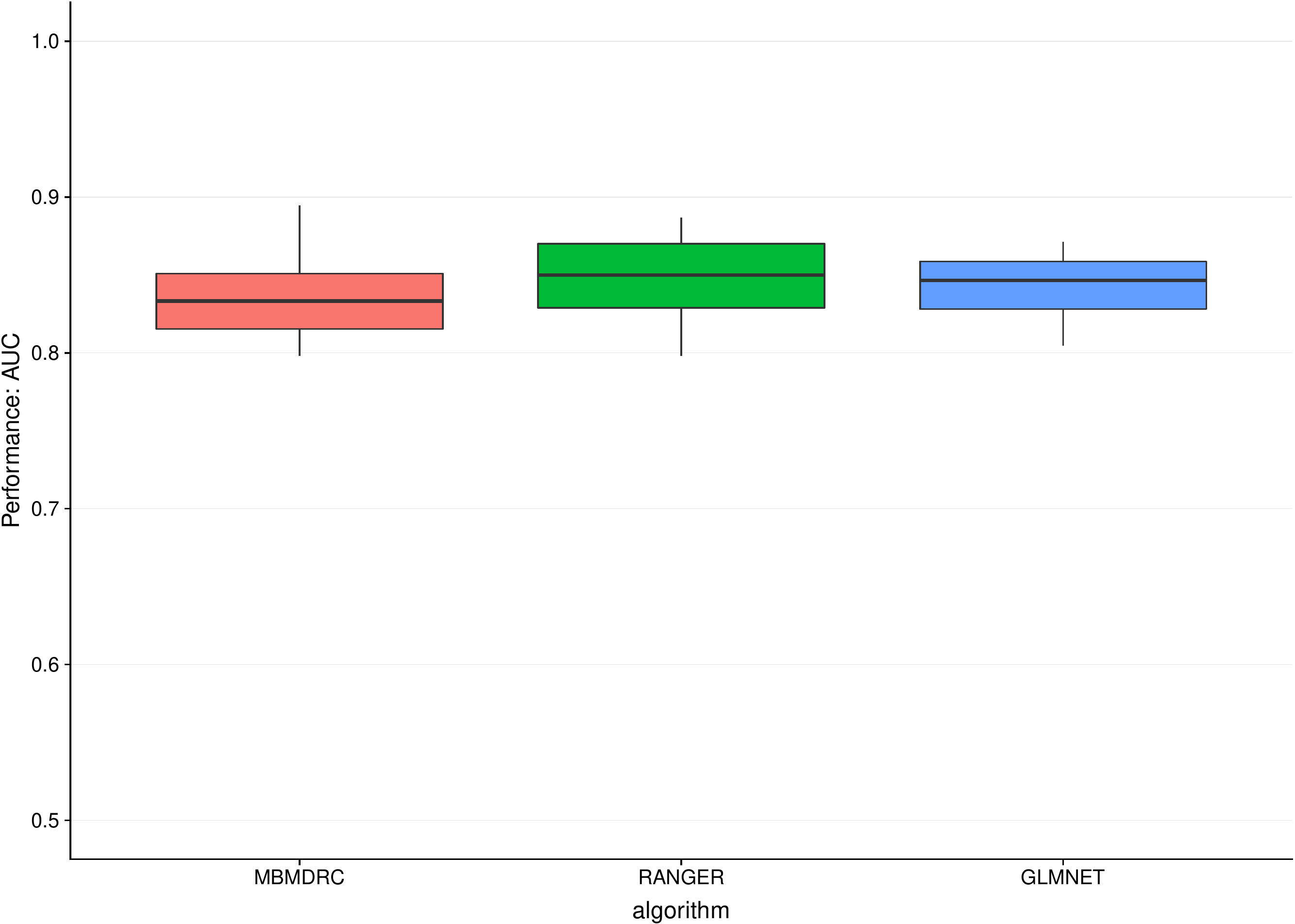} \caption{Performance in the NARAC dataset. Performance of the algorithms \texttt{MBMDRC}, \texttt{RANGER}, and \texttt{GLMNET} measured as AUC in 10 fold cross validation in the NARAC dataset.}\label{fig:narac-performance-plot}
\end{figure}

Application of the three algorithms to the rheumatoid arthritis dataset yields no relevant differences regarding their median performance.
Here, the \texttt{glmnet} models have a median AUC of 0.86, the \texttt{ranger} models of 0.85 and the \texttt{MBMDRC} models of 0.83, comparable to the simulation results of scenario 2, MAF=0.4, \(h^{2}=0.1\) and sample size between 1000 and 2000.
The corresponding box plot is shown in Figure \ref{fig:narac-performance-plot}.

Comparison of the runtimes shows that the \texttt{ranger} implementation is the fastest of all three.
Specifically, the mean runtime on Intel® Xeon® E5-2680 CPUs at 2.70GHz of one outer cross-validation fold is 1.5 times faster for \texttt{ranger} (378582.6 seconds) than for \texttt{MBMDRC} (603436.6 seconds), and 1.3 times faster for \texttt{glmnet} (465262.4 seconds) than for \texttt{MBMDRC}.

\FloatBarrier

\hypertarget{discussion}{%
\section{Discussion}\label{discussion}}

In this work, we extended a known algorithm to detect interactions to a classification algorithm that has a performance comparable to two popular classifications algorithms if no interactions are present but which clearly outperforms these if interactions are present.
We have shown this by a comprehensive simulation study and by application to real data.
Specifically, our simulation study revealed that our new classification algorithm can use information hidden in interactions more efficiently than the Random Forest approach, i.e.~smaller sample sizes are required to achieve similar performance.
The Elastic Net, at least in available implementations, does not consider interactions at all, thus is inappropriate if the outcome is influenced by interacting features.
In our application to the real dataset on RA, the performance of our algorithm was not relevantly different from that of the competitors, indicating that even though {[}\protect\hyperlink{ref-Liu2011}{41}{]} claimed to have identified putative interactions on chromosome 6 and MDR models of two SNPs entered the \texttt{MBMDRC} models, this did not improve the classification performance.

One drawback of our new method is the exponential increase in runtime with increasing the number of features in a dataset.
Whereas its runtime is not much slower than that of the Elastic Net approach, because both depend mostly on the number of features, Random Forest is clearly the fastest one, mainly dependent on the number of samples.
This makes the application of our new method to datasets on a genome-wide scale still challenging at the moment.

As a clear strength, we have shown in our unbiased benchmark on simulated data that taking interactions into account can improve classification performance.
As our method is not only applicable to biological/genetic data but to all datasets with discrete features, it may have practical implications in other applications, and we made our method available as an \texttt{R} package {[}\protect\hyperlink{ref-Gola2018}{33}{]}.

In addition, our observation that the Random Forest algorithm can make use of interacting features up to a certain degree fits well with Wright et al.~{[}\protect\hyperlink{ref-Wright2016}{19}{]}, who conclude that Random Forest is able to capture interactions but not to detect them.
In this regard, our method offers a clear advantage in that is not only able to capture interactions but the MDR models as the basic building block also allow insight into the underlying structure and dependencies among the features.

\FloatBarrier

\hypertarget{conclusions}{%
\section{Conclusions}\label{conclusions}}

We conclude that the explicit use of interactions between features can improve the prediction performance and thus should be included in further attempts to move precision medicine forward.

\FloatBarrier

\hypertarget{declarations}{%
\section{Declarations}\label{declarations}}

\hypertarget{acknowledgments}{%
\subsection{Acknowledgments}\label{acknowledgments}}

This work is based on data that was gathered with the support of grants from the National Institutes of Health (NO1-AR-2-2263 and RO1-AR-44422), and the National Arthritis Foundation.
We would like to thank Drs. Christopher I. Amos and Jean W. MacCluer, and Vanessa Olmo for the permission to use the dataset on rheumatoid arthritis.

\hypertarget{funding}{%
\subsection{Funding}\label{funding}}

This work was funded by the German Research Foundation (DFG, grant \#KO2240/-1 to IRK).

\hypertarget{availability-of-data-and-materials}{%
\subsection{Availability of data and materials}\label{availability-of-data-and-materials}}

All data generated during this study and the underlying \texttt{R} code are available on request.
The dataset on rheumatoid arthritis used during the current study is owned by the North American Rheumatoid Arthritis Consortium and is available from the corresponding author of {[}\protect\hyperlink{ref-Amos2009}{39}{]} on reasonable request.

\hypertarget{authors-contributions}{%
\subsection{Authors' contributions}\label{authors-contributions}}

DG designed and implemented the new algorithm, ran the simulations, analysed the NARAC dataset on rheumatoid arthritis, and drafted the manuscript.
IRK acquired the NARAC dataset and was involved in revising the manuscript critically.
All authors read and approved the final manuscript.

\hypertarget{ethics-approval-and-consent-to-participate}{%
\subsection{Ethics approval and consent to participate}\label{ethics-approval-and-consent-to-participate}}

Not applicable.

\hypertarget{consent-for-publication}{%
\subsection{Consent for publication}\label{consent-for-publication}}

Not applicable.

\hypertarget{competing-interests}{%
\subsection{Competing interests}\label{competing-interests}}

The authors declare that they have no competing interests.

\clearpage

\hypertarget{supplementary-information}{%
\section{Supplementary Information}\label{supplementary-information}}

\FloatBarrier

\FloatBarrier

\FloatBarrier

\FloatBarrier
\begin{table}[t]

\caption{\label{tab:performance-table-scenario1}Performance in scenario 1.}
\centering
\resizebox{\linewidth}{!}{
\begin{threeparttable}
\begin{tabular}{llrlll}
\toprule
MAF & $h^2$ & $n$ & MBMDRC & RANGER & GLMNET\\
\midrule
0.1 & 1 x 0.05 & 200 & 0.5391 (0.4940; 0.5858) & 0.5390 (0.5001; 0.5773) & 0.5316 (0.5000; 0.5854)\\
0.1 & 1 x 0.05 & 1000 & 0.5763 (0.5508; 0.5905) & 0.5855 (0.5716; 0.5952) & 0.5852 (0.5727; 0.5936)\\
0.1 & 1 x 0.05 & 2000 & 0.5825 (0.5678; 0.5934) & 0.5841 (0.5747; 0.6009) & 0.5880 (0.5781; 0.5934)\\
0.1 & 1 x 0.05 & 10000 & 0.5881 (0.5831; 0.5926) & 0.5874 (0.5813; 0.5923) & 0.5883 (0.5836; 0.5912)\\
0.1 & 1 x 0.1 & 200 & 0.5863 (0.5300; 0.6391) & 0.6142 (0.5665; 0.6423) & 0.6000 (0.5000; 0.6314)\\
\addlinespace
0.1 & 1 x 0.1 & 1000 & 0.6117 (0.5927; 0.6355) & 0.6215 (0.6049; 0.6328) & 0.6129 (0.6053; 0.6297)\\
0.1 & 1 x 0.1 & 2000 & 0.6196 (0.6098; 0.6316) & 0.6239 (0.6102; 0.6342) & 0.6233 (0.6153; 0.6314)\\
0.1 & 1 x 0.1 & 10000 & 0.6233 (0.6206; 0.6279) & 0.6239 (0.6195; 0.6279) & 0.6246 (0.6207; 0.6278)\\
0.1 & 1 x 0.2 & 200 & 0.6503 (0.6041; 0.7003) & 0.6800 (0.6544; 0.7046) & 0.6702 (0.6408; 0.7011)\\
0.1 & 1 x 0.2 & 1000 & 0.6737 (0.6585; 0.6859) & 0.6771 (0.6654; 0.6929) & 0.6750 (0.6682; 0.6830)\\
\addlinespace
0.1 & 1 x 0.2 & 2000 & 0.6765 (0.6664; 0.6875) & 0.6772 (0.6686; 0.6910) & 0.6801 (0.6674; 0.6908)\\
0.1 & 1 x 0.2 & 10000 & 0.6814 (0.6772; 0.6864) & 0.6813 (0.6764; 0.6859) & 0.6817 (0.6763; 0.6842)\\
0.2 & 1 x 0.05 & 200 & 0.5321 (0.4633; 0.5679) & 0.5454 (0.4958; 0.5774) & 0.5000 (0.5000; 0.5725)\\
0.2 & 1 x 0.05 & 1000 & 0.5940 (0.5697; 0.6155) & 0.5981 (0.5791; 0.6148) & 0.6040 (0.5904; 0.6215)\\
0.2 & 1 x 0.05 & 2000 & 0.6022 (0.5905; 0.6197) & 0.6080 (0.5946; 0.6180) & 0.6104 (0.6013; 0.6220)\\
\addlinespace
0.2 & 1 x 0.05 & 10000 & 0.6103 (0.6063; 0.6138) & 0.6099 (0.6068; 0.6137) & 0.6095 (0.6058; 0.6130)\\
0.2 & 1 x 0.1 & 200 & 0.6021 (0.5410; 0.6523) & 0.6216 (0.5665; 0.6706) & 0.6144 (0.5000; 0.6646)\\
0.2 & 1 x 0.1 & 1000 & 0.6529 (0.6314; 0.6703) & 0.6581 (0.6352; 0.6747) & 0.6580 (0.6445; 0.6706)\\
0.2 & 1 x 0.1 & 2000 & 0.6536 (0.6390; 0.6713) & 0.6536 (0.6416; 0.6650) & 0.6582 (0.6457; 0.6657)\\
0.2 & 1 x 0.1 & 10000 & 0.6551 (0.6513; 0.6609) & 0.6558 (0.6494; 0.6596) & 0.6556 (0.6509; 0.6587)\\
\addlinespace
0.2 & 1 x 0.2 & 200 & 0.6901 (0.6580; 0.7326) & 0.7116 (0.6757; 0.7440) & 0.7150 (0.6913; 0.7418)\\
0.2 & 1 x 0.2 & 1000 & 0.7146 (0.6931; 0.7288) & 0.7174 (0.7040; 0.7268) & 0.7193 (0.7033; 0.7294)\\
0.2 & 1 x 0.2 & 2000 & 0.7163 (0.7031; 0.7273) & 0.7150 (0.7076; 0.7273) & 0.7190 (0.7105; 0.7307)\\
0.2 & 1 x 0.2 & 10000 & 0.7179 (0.7123; 0.7230) & 0.7167 (0.7113; 0.7222) & 0.7186 (0.7134; 0.7224)\\
0.4 & 1 x 0.05 & 200 & 0.5234 (0.4751; 0.5776) & 0.5460 (0.4984; 0.5871) & 0.5000 (0.5000; 0.5844)\\
\addlinespace
0.4 & 1 x 0.05 & 1000 & 0.6059 (0.5879; 0.6280) & 0.6038 (0.5902; 0.6202) & 0.6190 (0.6037; 0.6339)\\
0.4 & 1 x 0.05 & 2000 & 0.6172 (0.6027; 0.6301) & 0.6145 (0.5981; 0.6295) & 0.6212 (0.6088; 0.6349)\\
0.4 & 1 x 0.05 & 10000 & 0.6249 (0.6182; 0.6293) & 0.6235 (0.6180; 0.6275) & 0.6261 (0.6192; 0.6293)\\
0.4 & 1 x 0.1 & 200 & 0.6046 (0.5518; 0.6644) & 0.6350 (0.5636; 0.6722) & 0.6428 (0.5554; 0.6880)\\
0.4 & 1 x 0.1 & 1000 & 0.6743 (0.6497; 0.6846) & 0.6724 (0.6504; 0.6852) & 0.6799 (0.6677; 0.6891)\\
\addlinespace
0.4 & 1 x 0.1 & 2000 & 0.6799 (0.6650; 0.6916) & 0.6735 (0.6643; 0.6853) & 0.6802 (0.6706; 0.6895)\\
0.4 & 1 x 0.1 & 10000 & 0.6765 (0.6729; 0.6831) & 0.6767 (0.6713; 0.6826) & 0.6769 (0.6731; 0.6819)\\
0.4 & 1 x 0.2 & 200 & 0.6952 (0.6368; 0.7480) & 0.7202 (0.6901; 0.7531) & 0.7468 (0.7120; 0.7661)\\
0.4 & 1 x 0.2 & 1000 & 0.7337 (0.7115; 0.7546) & 0.7323 (0.7150; 0.7520) & 0.7420 (0.7242; 0.7589)\\
0.4 & 1 x 0.2 & 2000 & 0.7476 (0.7343; 0.7570) & 0.7445 (0.7357; 0.7573) & 0.7527 (0.7433; 0.7615)\\
\addlinespace
0.4 & 1 x 0.2 & 10000 & 0.7487 (0.7430; 0.7523) & 0.7488 (0.7417; 0.7526) & 0.7480 (0.7435; 0.7520)\\
\bottomrule
\end{tabular}
\begin{tablenotes}
\item \textit{Note: } 
\item Performance of the algorithms \texttt{MBMDRC}, \texttt{RANGER}, and \texttt{GLMNET} measured as AUC over 50 replicates in scenario 7. The median of the AUC and the 25
\end{tablenotes}
\end{threeparttable}}
\end{table}

\FloatBarrier
\begin{table}[t]

\caption{\label{tab:performance-table-scenario2}Performance in scenario 2.}
\centering
\resizebox{\linewidth}{!}{
\begin{threeparttable}
\begin{tabular}{llrlll}
\toprule
MAF & $h^2$ & $n$ & MBMDRC & RANGER & GLMNET\\
\midrule
0.1 & 5 x 0.05 & 200 & 0.5504 (0.4952; 0.6008) & 0.6032 (0.5427; 0.6428) & 0.5608 (0.5000; 0.6338)\\
0.1 & 5 x 0.05 & 1000 & 0.7132 (0.6941; 0.7381) & 0.7183 (0.6984; 0.7337) & 0.7166 (0.6933; 0.7356)\\
0.1 & 5 x 0.05 & 2000 & 0.7338 (0.7221; 0.7437) & 0.7248 (0.7159; 0.7383) & 0.7280 (0.7170; 0.7432)\\
0.1 & 5 x 0.05 & 10000 & 0.7361 (0.7306; 0.7397) & 0.7323 (0.7262; 0.7361) & 0.7343 (0.7301; 0.7393)\\
0.1 & 5 x 0.1 & 200 & 0.6400 (0.5706; 0.7052) & 0.6990 (0.6516; 0.7354) & 0.6558 (0.5936; 0.7220)\\
\addlinespace
0.1 & 5 x 0.1 & 1000 & 0.8098 (0.7952; 0.8277) & 0.7965 (0.7761; 0.8168) & 0.8012 (0.7841; 0.8234)\\
0.1 & 5 x 0.1 & 2000 & 0.8101 (0.8022; 0.8191) & 0.8037 (0.7943; 0.8110) & 0.8077 (0.8002; 0.8156)\\
0.1 & 5 x 0.1 & 10000 & 0.8164 (0.8109; 0.8197) & 0.8127 (0.8096; 0.8165) & 0.8160 (0.8113; 0.8193)\\
0.1 & 5 x 0.2 & 200 & 0.7818 (0.7249; 0.8383) & 0.8032 (0.7649; 0.8503) & 0.8092 (0.7656; 0.8645)\\
0.1 & 5 x 0.2 & 1000 & 0.8907 (0.8817; 0.9011) & 0.8830 (0.8732; 0.8915) & 0.8920 (0.8817; 0.9006)\\
\addlinespace
0.1 & 5 x 0.2 & 2000 & 0.8929 (0.8862; 0.8974) & 0.8848 (0.8796; 0.8915) & 0.8916 (0.8866; 0.8977)\\
0.1 & 5 x 0.2 & 10000 & 0.8969 (0.8944; 0.8999) & 0.8936 (0.8916; 0.8957) & 0.8947 (0.8916; 0.8981)\\
0.2 & 5 x 0.05 & 200 & 0.5786 (0.5248; 0.6410) & 0.6094 (0.5603; 0.6765) & 0.5880 (0.5000; 0.6302)\\
0.2 & 5 x 0.05 & 1000 & 0.7303 (0.7053; 0.7520) & 0.7134 (0.6944; 0.7341) & 0.7311 (0.7041; 0.7517)\\
0.2 & 5 x 0.05 & 2000 & 0.7421 (0.7257; 0.7562) & 0.7258 (0.7125; 0.7389) & 0.7357 (0.7218; 0.7475)\\
\addlinespace
0.2 & 5 x 0.05 & 10000 & 0.7457 (0.7412; 0.7501) & 0.7402 (0.7340; 0.7445) & 0.7444 (0.7387; 0.7490)\\
0.2 & 5 x 0.1 & 200 & 0.6816 (0.6122; 0.7511) & 0.7038 (0.6554; 0.7563) & 0.6934 (0.6345; 0.7425)\\
0.2 & 5 x 0.1 & 1000 & 0.8146 (0.7990; 0.8233) & 0.7969 (0.7791; 0.8044) & 0.8019 (0.7898; 0.8193)\\
0.2 & 5 x 0.1 & 2000 & 0.8105 (0.8018; 0.8221) & 0.7985 (0.7926; 0.8095) & 0.8078 (0.8002; 0.8188)\\
0.2 & 5 x 0.1 & 10000 & 0.8140 (0.8098; 0.8177) & 0.8092 (0.8037; 0.8132) & 0.8131 (0.8077; 0.8164)\\
\addlinespace
0.2 & 5 x 0.2 & 200 & 0.7663 (0.7160; 0.8126) & 0.7884 (0.7371; 0.8228) & 0.8061 (0.7320; 0.8504)\\
0.2 & 5 x 0.2 & 1000 & 0.8772 (0.8655; 0.8870) & 0.8595 (0.8497; 0.8741) & 0.8744 (0.8602; 0.8858)\\
0.2 & 5 x 0.2 & 2000 & 0.8777 (0.8718; 0.8860) & 0.8675 (0.8608; 0.8741) & 0.8755 (0.8673; 0.8818)\\
0.2 & 5 x 0.2 & 10000 & 0.8814 (0.8786; 0.8845) & 0.8761 (0.8736; 0.8788) & 0.8785 (0.8757; 0.8817)\\
0.4 & 5 x 0.05 & 200 & 0.5946 (0.5553; 0.6322) & 0.6424 (0.5873; 0.6855) & 0.6445 (0.5635; 0.6894)\\
\addlinespace
0.4 & 5 x 0.05 & 1000 & 0.7510 (0.7273; 0.7660) & 0.7323 (0.7162; 0.7458) & 0.7504 (0.7269; 0.7624)\\
0.4 & 5 x 0.05 & 2000 & 0.7609 (0.7499; 0.7726) & 0.7449 (0.7367; 0.7528) & 0.7560 (0.7475; 0.7681)\\
0.4 & 5 x 0.05 & 10000 & 0.7670 (0.7615; 0.7713) & 0.7591 (0.7532; 0.7642) & 0.7638 (0.7586; 0.7694)\\
0.4 & 5 x 0.1 & 200 & 0.6922 (0.6408; 0.7463) & 0.7362 (0.6795; 0.7702) & 0.7177 (0.6828; 0.7872)\\
0.4 & 5 x 0.1 & 1000 & 0.8449 (0.8278; 0.8583) & 0.8257 (0.8059; 0.8377) & 0.8384 (0.8243; 0.8518)\\
\addlinespace
0.4 & 5 x 0.1 & 2000 & 0.8446 (0.8372; 0.8533) & 0.8298 (0.8231; 0.8380) & 0.8393 (0.8317; 0.8471)\\
0.4 & 5 x 0.1 & 10000 & 0.8474 (0.8424; 0.8502) & 0.8412 (0.8368; 0.8440) & 0.8459 (0.8422; 0.8490)\\
0.4 & 5 x 0.2 & 200 & 0.8122 (0.7662; 0.8564) & 0.8390 (0.8094; 0.8668) & 0.8415 (0.7967; 0.8766)\\
0.4 & 5 x 0.2 & 1000 & 0.9177 (0.9077; 0.9269) & 0.9003 (0.8935; 0.9118) & 0.9121 (0.9026; 0.9213)\\
0.4 & 5 x 0.2 & 2000 & 0.9169 (0.9124; 0.9223) & 0.9070 (0.9027; 0.9111) & 0.9130 (0.9082; 0.9176)\\
\addlinespace
0.4 & 5 x 0.2 & 10000 & 0.9166 (0.9141; 0.9189) & 0.9122 (0.9095; 0.9149) & 0.9129 (0.9110; 0.9155)\\
\bottomrule
\end{tabular}
\begin{tablenotes}
\item \textit{Note: } 
\item Performance of the algorithms \texttt{MBMDRC}, \texttt{RANGER}, and \texttt{GLMNET} measured as AUC over 50 replicates in scenario 7. The median of the AUC and the 25
\end{tablenotes}
\end{threeparttable}}
\end{table}

\FloatBarrier

\begin{ThreePartTable}
\begin{TableNotes}
\item \textit{Note: } 
\item Performance of the algorithms \texttt{MBMDRC}, \texttt{RANGER}, and \texttt{GLMNET} measured as AUC over 50 replicates in scenario 7. The median of the AUC and the 25
\end{TableNotes}
\begin{longtable}{llrlll}
\caption{\label{tab:performance-table-scenario3}Performance in scenario 3.}\\
\toprule
MAF & $h^2$ & $n$ & MBMDRC & RANGER & GLMNET\\
\midrule
\endfirsthead
\caption[]{\label{tab:performance-table-scenario3}Performance in scenario 3. \textit{(continued)}}\\
\toprule
MAF & $h^2$ & $n$ & MBMDRC & RANGER & GLMNET\\
\midrule
\endhead
\
\endfoot
\bottomrule
\insertTableNotes
\endlastfoot
0.1,0.1 & 1 x 0.05 & 200 & 0.5355 (0.4856; 0.5861) & 0.4996 (0.4543; 0.5375) & 0.5000 (0.5000; 0.5268)\\
0.1,0.1 & 1 x 0.05 & 1000 & 0.5430 (0.5101; 0.5836) & 0.5017 (0.4837; 0.5189) & 0.5000 (0.4961; 0.5036)\\
0.1,0.1 & 1 x 0.05 & 2000 & 0.5796 (0.5647; 0.5929) & 0.5018 (0.4866; 0.5139) & 0.5000 (0.4960; 0.5013)\\
0.1,0.1 & 1 x 0.05 & 10000 & 0.5886 (0.5835; 0.5950) & 0.5196 (0.5143; 0.5283) & 0.5000 (0.4972; 0.5000)\\
0.1,0.1 & 1 x 0.1 & 200 & 0.5020 (0.4685; 0.5432) & 0.4914 (0.4557; 0.5154) & 0.5000 (0.4873; 0.5000)\\
\addlinespace
0.1,0.1 & 1 x 0.1 & 1000 & 0.6523 (0.6397; 0.6703) & 0.5138 (0.4987; 0.5377) & 0.5000 (0.4984; 0.5007)\\
0.1,0.1 & 1 x 0.1 & 2000 & 0.6649 (0.6566; 0.6767) & 0.5194 (0.5045; 0.5375) & 0.5000 (0.4937; 0.5000)\\
0.1,0.1 & 1 x 0.1 & 10000 & 0.6671 (0.6587; 0.6749) & 0.5601 (0.5515; 0.5735) & 0.5000 (0.4980; 0.5015)\\
0.1,0.1 & 1 x 0.2 & 200 & 0.5602 (0.4945; 0.6581) & 0.4934 (0.4436; 0.5268) & 0.5000 (0.4702; 0.5000)\\
0.1,0.1 & 1 x 0.2 & 1000 & 0.7248 (0.7113; 0.7352) & 0.5326 (0.5089; 0.5572) & 0.5000 (0.5000; 0.5128)\\
\addlinespace
0.1,0.1 & 1 x 0.2 & 2000 & 0.7230 (0.7168; 0.7318) & 0.5507 (0.5270; 0.5787) & 0.5000 (0.4851; 0.5000)\\
0.1,0.1 & 1 x 0.2 & 10000 & 0.7253 (0.7210; 0.7293) & 0.6316 (0.6142; 0.6507) & 0.5000 (0.5000; 0.5033)\\
0.1,0.2 & 1 x 0.05 & 200 & 0.5100 (0.4689; 0.5698) & 0.4946 (0.4587; 0.5557) & 0.5000 (0.4814; 0.5158)\\
0.1,0.2 & 1 x 0.05 & 1000 & 0.5883 (0.5271; 0.6152) & 0.5282 (0.5036; 0.5472) & 0.5000 (0.5000; 0.5201)\\
0.1,0.2 & 1 x 0.05 & 2000 & 0.6178 (0.6069; 0.6287) & 0.5555 (0.5237; 0.5798) & 0.5000 (0.5000; 0.5261)\\
\addlinespace
0.1,0.2 & 1 x 0.05 & 10000 & 0.6249 (0.6168; 0.6348) & 0.6201 (0.6119; 0.6314) & 0.5396 (0.5290; 0.5619)\\
0.1,0.2 & 1 x 0.1 & 200 & 0.5180 (0.4806; 0.5636) & 0.5046 (0.4530; 0.5420) & 0.5000 (0.5000; 0.5258)\\
0.1,0.2 & 1 x 0.1 & 1000 & 0.6678 (0.6486; 0.6880) & 0.5789 (0.5329; 0.6176) & 0.5106 (0.5000; 0.5405)\\
0.1,0.2 & 1 x 0.1 & 2000 & 0.6701 (0.6618; 0.6798) & 0.6483 (0.6285; 0.6592) & 0.5336 (0.5000; 0.5533)\\
0.1,0.2 & 1 x 0.1 & 10000 & 0.6735 (0.6680; 0.6822) & 0.6728 (0.6658; 0.6806) & 0.5962 (0.5635; 0.6042)\\
\addlinespace
0.1,0.2 & 1 x 0.2 & 200 & 0.6561 (0.5819; 0.7335) & 0.5410 (0.4758; 0.6090) & 0.5000 (0.4933; 0.5448)\\
0.1,0.2 & 1 x 0.2 & 1000 & 0.7685 (0.7556; 0.7814) & 0.7514 (0.7326; 0.7740) & 0.5821 (0.5509; 0.6194)\\
0.1,0.2 & 1 x 0.2 & 2000 & 0.7692 (0.7619; 0.7759) & 0.7695 (0.7604; 0.7781) & 0.6064 (0.5807; 0.6561)\\
0.1,0.2 & 1 x 0.2 & 10000 & 0.7753 (0.7690; 0.7795) & 0.7750 (0.7709; 0.7800) & 0.6654 (0.6556; 0.6761)\\
0.1,0.4 & 1 x 0.05 & 200 & 0.5533 (0.5023; 0.6112) & 0.5204 (0.4889; 0.5752) & 0.5000 (0.4884; 0.5300)\\
\addlinespace
0.1,0.4 & 1 x 0.05 & 1000 & 0.6268 (0.5383; 0.6830) & 0.5767 (0.5305; 0.6662) & 0.5214 (0.5000; 0.5548)\\
0.1,0.4 & 1 x 0.05 & 2000 & 0.6520 (0.5901; 0.6871) & 0.6315 (0.5621; 0.6824) & 0.5443 (0.5005; 0.5982)\\
0.1,0.4 & 1 x 0.05 & 10000 & 0.6643 (0.6044; 0.6912) & 0.6554 (0.5951; 0.6875) & 0.5595 (0.5370; 0.6192)\\
0.1,0.4 & 1 x 0.1 & 200 & 0.6928 (0.5658; 0.7620) & 0.5996 (0.5443; 0.6825) & 0.5000 (0.4941; 0.5454)\\
0.1,0.4 & 1 x 0.1 & 1000 & 0.7665 (0.7095; 0.8050) & 0.7559 (0.6752; 0.7977) & 0.5700 (0.5343; 0.6364)\\
\addlinespace
0.1,0.4 & 1 x 0.1 & 2000 & 0.7903 (0.6991; 0.8132) & 0.7776 (0.6919; 0.8066) & 0.5900 (0.5570; 0.6695)\\
0.1,0.4 & 1 x 0.1 & 10000 & 0.7965 (0.7081; 0.8158) & 0.7892 (0.7071; 0.8119) & 0.6205 (0.5674; 0.6738)\\
0.1,0.4 & 1 x 0.2 & 200 & 0.8393 (0.8002; 0.8731) & 0.7616 (0.6871; 0.8072) & 0.5478 (0.5000; 0.5964)\\
0.1,0.4 & 1 x 0.2 & 1000 & 0.8626 (0.8381; 0.8793) & 0.8427 (0.8320; 0.8665) & 0.6039 (0.5832; 0.6559)\\
0.1,0.4 & 1 x 0.2 & 2000 & 0.8540 (0.8460; 0.8846) & 0.8555 (0.8426; 0.8745) & 0.6597 (0.6041; 0.7034)\\
\addlinespace
0.1,0.4 & 1 x 0.2 & 10000 & 0.8522 (0.8447; 0.8810) & 0.8516 (0.8421; 0.8774) & 0.6703 (0.6342; 0.7087)\\
0.2,0.2 & 1 x 0.05 & 200 & 0.5036 (0.4561; 0.5484) & 0.4944 (0.4529; 0.5404) & 0.5000 (0.4896; 0.5173)\\
0.2,0.2 & 1 x 0.05 & 1000 & 0.5494 (0.5054; 0.5927) & 0.5007 (0.4851; 0.5155) & 0.5000 (0.4951; 0.5048)\\
0.2,0.2 & 1 x 0.05 & 2000 & 0.6084 (0.5931; 0.6228) & 0.4982 (0.4867; 0.5133) & 0.5000 (0.4921; 0.5000)\\
0.2,0.2 & 1 x 0.05 & 10000 & 0.6146 (0.6101; 0.6206) & 0.5192 (0.5109; 0.5317) & 0.5000 (0.4967; 0.5000)\\
\addlinespace
0.2,0.2 & 1 x 0.1 & 200 & 0.5116 (0.4673; 0.5753) & 0.4938 (0.4588; 0.5414) & 0.5000 (0.4824; 0.5070)\\
0.2,0.2 & 1 x 0.1 & 1000 & 0.6568 (0.6373; 0.6721) & 0.5029 (0.4874; 0.5209) & 0.5000 (0.4961; 0.5030)\\
0.2,0.2 & 1 x 0.1 & 2000 & 0.6637 (0.6498; 0.6731) & 0.5092 (0.4949; 0.5275) & 0.5000 (0.4919; 0.5037)\\
0.2,0.2 & 1 x 0.1 & 10000 & 0.6644 (0.6605; 0.6685) & 0.5434 (0.5310; 0.5678) & 0.5000 (0.4967; 0.5000)\\
0.2,0.2 & 1 x 0.2 & 200 & 0.5590 (0.5086; 0.6306) & 0.4966 (0.4506; 0.5392) & 0.5000 (0.5000; 0.5136)\\
\addlinespace
0.2,0.2 & 1 x 0.2 & 1000 & 0.7264 (0.7098; 0.7381) & 0.5164 (0.4948; 0.5376) & 0.5000 (0.4920; 0.5047)\\
0.2,0.2 & 1 x 0.2 & 2000 & 0.7285 (0.7168; 0.7442) & 0.5343 (0.5132; 0.5570) & 0.5000 (0.4935; 0.5000)\\
0.2,0.2 & 1 x 0.2 & 10000 & 0.7326 (0.7279; 0.7389) & 0.5983 (0.5839; 0.6260) & 0.5000 (0.5000; 0.5000)\\
0.2,0.4 & 1 x 0.05 & 200 & 0.5276 (0.4668; 0.5744) & 0.5098 (0.4609; 0.5491) & 0.5000 (0.5000; 0.5328)\\
0.2,0.4 & 1 x 0.05 & 1000 & 0.5835 (0.5269; 0.6374) & 0.5328 (0.5086; 0.5781) & 0.5013 (0.5000; 0.5391)\\
\addlinespace
0.2,0.4 & 1 x 0.05 & 2000 & 0.6311 (0.5739; 0.6651) & 0.5808 (0.5075; 0.6223) & 0.5080 (0.5000; 0.5610)\\
0.2,0.4 & 1 x 0.05 & 10000 & 0.6398 (0.5932; 0.6544) & 0.6282 (0.5554; 0.6552) & 0.5261 (0.5000; 0.5855)\\
0.2,0.4 & 1 x 0.1 & 200 & 0.5256 (0.4872; 0.5790) & 0.5104 (0.4678; 0.5663) & 0.5000 (0.4922; 0.5268)\\
0.2,0.4 & 1 x 0.1 & 1000 & 0.6491 (0.5827; 0.7144) & 0.5839 (0.5063; 0.6538) & 0.5000 (0.4978; 0.5203)\\
0.2,0.4 & 1 x 0.1 & 2000 & 0.6742 (0.6370; 0.7206) & 0.6481 (0.5373; 0.7022) & 0.5044 (0.5000; 0.5515)\\
\addlinespace
0.2,0.4 & 1 x 0.1 & 10000 & 0.6879 (0.6431; 0.7219) & 0.6822 (0.6304; 0.7176) & 0.5330 (0.5068; 0.5913)\\
0.2,0.4 & 1 x 0.2 & 200 & 0.7010 (0.5708; 0.7922) & 0.5684 (0.5155; 0.6415) & 0.5147 (0.5000; 0.5636)\\
0.2,0.4 & 1 x 0.2 & 1000 & 0.7787 (0.7506; 0.8096) & 0.7704 (0.6835; 0.8070) & 0.5463 (0.5001; 0.6314)\\
0.2,0.4 & 1 x 0.2 & 2000 & 0.7932 (0.7400; 0.8113) & 0.7866 (0.7200; 0.7985) & 0.5711 (0.5292; 0.6756)\\
0.2,0.4 & 1 x 0.2 & 10000 & 0.7964 (0.7518; 0.8049) & 0.7946 (0.7512; 0.8034) & 0.6152 (0.5583; 0.6692)\\
\addlinespace
0.4,0.4 & 1 x 0.05 & 200 & 0.5189 (0.4813; 0.5589) & 0.4928 (0.4548; 0.5414) & 0.5000 (0.4896; 0.5112)\\
0.4,0.4 & 1 x 0.05 & 1000 & 0.5472 (0.5195; 0.5941) & 0.4983 (0.4830; 0.5140) & 0.5000 (0.4937; 0.5015)\\
0.4,0.4 & 1 x 0.05 & 2000 & 0.6138 (0.5950; 0.6294) & 0.4999 (0.4889; 0.5143) & 0.5000 (0.5000; 0.5027)\\
0.4,0.4 & 1 x 0.05 & 10000 & 0.6206 (0.6170; 0.6260) & 0.5165 (0.5100; 0.5265) & 0.5000 (0.5000; 0.5000)\\
0.4,0.4 & 1 x 0.1 & 200 & 0.5176 (0.4690; 0.5543) & 0.4828 (0.4570; 0.5369) & 0.5000 (0.4943; 0.5260)\\
\addlinespace
0.4,0.4 & 1 x 0.1 & 1000 & 0.6662 (0.6480; 0.6888) & 0.5041 (0.4872; 0.5225) & 0.5000 (0.4952; 0.5046)\\
0.4,0.4 & 1 x 0.1 & 2000 & 0.6708 (0.6606; 0.6804) & 0.5073 (0.4931; 0.5241) & 0.5000 (0.4918; 0.5003)\\
0.4,0.4 & 1 x 0.1 & 10000 & 0.6730 (0.6656; 0.6797) & 0.5472 (0.5362; 0.5626) & 0.5000 (0.4972; 0.5000)\\
0.4,0.4 & 1 x 0.2 & 200 & 0.5666 (0.5070; 0.6566) & 0.5026 (0.4508; 0.5461) & 0.5000 (0.4864; 0.5258)\\
0.4,0.4 & 1 x 0.2 & 1000 & 0.7412 (0.7312; 0.7508) & 0.5133 (0.4943; 0.5280) & 0.5000 (0.4965; 0.5008)\\
\addlinespace
0.4,0.4 & 1 x 0.2 & 2000 & 0.7477 (0.7371; 0.7591) & 0.5338 (0.5135; 0.5514) & 0.5000 (0.4930; 0.5029)\\
0.4,0.4 & 1 x 0.2 & 10000 & 0.7479 (0.7414; 0.7528) & 0.6198 (0.5903; 0.6565) & 0.5000 (0.4968; 0.5000)\\*
\end{longtable}
\end{ThreePartTable}

\FloatBarrier

\begin{ThreePartTable}
\begin{TableNotes}
\item \textit{Note: } 
\item Performance of the algorithms \texttt{MBMDRC}, \texttt{RANGER}, and \texttt{GLMNET} measured as AUC over 50 replicates in scenario 7. The median of the AUC and the 25
\end{TableNotes}
\begin{longtable}{llrlll}
\caption{\label{tab:performance-table-scenario4}Performance in scenario 4.}\\
\toprule
MAF & $h^2$ & $n$ & MBMDRC & RANGER & GLMNET\\
\midrule
\endfirsthead
\caption[]{\label{tab:performance-table-scenario4}Performance in scenario 4. \textit{(continued)}}\\
\toprule
MAF & $h^2$ & $n$ & MBMDRC & RANGER & GLMNET\\
\midrule
\endhead
\
\endfoot
\bottomrule
\insertTableNotes
\endlastfoot
0.1,0.1; 0.1; 0.2; 0.4 & 4 x 0.05 & 200 & 0.5415 (0.4867; 0.6123) & 0.5702 (0.5132; 0.6059) & 0.5381 (0.5000; 0.5929)\\
0.1,0.1; 0.1; 0.2; 0.4 & 4 x 0.05 & 1000 & 0.6729 (0.6467; 0.6989) & 0.6593 (0.6444; 0.6807) & 0.6721 (0.6402; 0.6951)\\
0.1,0.1; 0.1; 0.2; 0.4 & 4 x 0.05 & 2000 & 0.7120 (0.6948; 0.7316) & 0.6813 (0.6709; 0.6968) & 0.6812 (0.6667; 0.6956)\\
0.1,0.1; 0.1; 0.2; 0.4 & 4 x 0.05 & 10000 & 0.7311 (0.7234; 0.7400) & 0.7049 (0.6949; 0.7203) & 0.6903 (0.6794; 0.6993)\\
0.1,0.1; 0.1; 0.2; 0.4 & 4 x 0.1 & 200 & 0.5910 (0.5397; 0.6520) & 0.6496 (0.5891; 0.6943) & 0.6474 (0.5801; 0.6945)\\
\addlinespace
0.1,0.1; 0.1; 0.2; 0.4 & 4 x 0.1 & 1000 & 0.7910 (0.7737; 0.8082) & 0.7356 (0.7146; 0.7492) & 0.7339 (0.7128; 0.7587)\\
0.1,0.1; 0.1; 0.2; 0.4 & 4 x 0.1 & 2000 & 0.8102 (0.7947; 0.8239) & 0.7484 (0.7338; 0.7629) & 0.7460 (0.7327; 0.7575)\\
0.1,0.1; 0.1; 0.2; 0.4 & 4 x 0.1 & 10000 & 0.8170 (0.8112; 0.8215) & 0.7792 (0.7643; 0.7951) & 0.7545 (0.7446; 0.7666)\\
0.1,0.1; 0.1; 0.2; 0.4 & 4 x 0.2 & 200 & 0.6695 (0.6109; 0.7443) & 0.7000 (0.6563; 0.7300) & 0.7099 (0.6570; 0.7498)\\
0.1,0.1; 0.1; 0.2; 0.4 & 4 x 0.2 & 1000 & 0.8804 (0.8666; 0.8951) & 0.7982 (0.7845; 0.8179) & 0.7706 (0.7565; 0.7894)\\
\addlinespace
0.1,0.1; 0.1; 0.2; 0.4 & 4 x 0.2 & 2000 & 0.8907 (0.8819; 0.8991) & 0.8279 (0.8171; 0.8427) & 0.7823 (0.7693; 0.7952)\\
0.1,0.1; 0.1; 0.2; 0.4 & 4 x 0.2 & 10000 & 0.8938 (0.8899; 0.8974) & 0.8756 (0.8724; 0.8797) & 0.7968 (0.7840; 0.8081)\\
0.1,0.2; 0.1; 0.2; 0.4 & 4 x 0.05 & 200 & 0.5612 (0.5042; 0.6228) & 0.5844 (0.5241; 0.6302) & 0.5520 (0.5000; 0.6397)\\
0.1,0.2; 0.1; 0.2; 0.4 & 4 x 0.05 & 1000 & 0.6700 (0.6440; 0.6997) & 0.6676 (0.6541; 0.6902) & 0.6639 (0.6480; 0.6888)\\
0.1,0.2; 0.1; 0.2; 0.4 & 4 x 0.05 & 2000 & 0.7205 (0.7057; 0.7345) & 0.6819 (0.6697; 0.6914) & 0.6819 (0.6701; 0.6938)\\
\addlinespace
0.1,0.2; 0.1; 0.2; 0.4 & 4 x 0.05 & 10000 & 0.7385 (0.7308; 0.7439) & 0.7134 (0.7097; 0.7194) & 0.6963 (0.6918; 0.7013)\\
0.1,0.2; 0.1; 0.2; 0.4 & 4 x 0.1 & 200 & 0.5800 (0.5299; 0.6376) & 0.6242 (0.5934; 0.6695) & 0.6096 (0.5438; 0.6592)\\
0.1,0.2; 0.1; 0.2; 0.4 & 4 x 0.1 & 1000 & 0.7678 (0.7406; 0.7917) & 0.7369 (0.7173; 0.7541) & 0.7324 (0.7077; 0.7439)\\
0.1,0.2; 0.1; 0.2; 0.4 & 4 x 0.1 & 2000 & 0.8004 (0.7911; 0.8088) & 0.7541 (0.7373; 0.7640) & 0.7372 (0.7235; 0.7511)\\
0.1,0.2; 0.1; 0.2; 0.4 & 4 x 0.1 & 10000 & 0.8141 (0.8094; 0.8169) & 0.7961 (0.7914; 0.8008) & 0.7551 (0.7420; 0.7618)\\
\addlinespace
0.1,0.2; 0.1; 0.2; 0.4 & 4 x 0.2 & 200 & 0.6662 (0.6113; 0.7121) & 0.6990 (0.6601; 0.7396) & 0.6706 (0.6248; 0.7082)\\
0.1,0.2; 0.1; 0.2; 0.4 & 4 x 0.2 & 1000 & 0.8718 (0.8469; 0.8897) & 0.8458 (0.8275; 0.8641) & 0.7793 (0.7616; 0.7960)\\
0.1,0.2; 0.1; 0.2; 0.4 & 4 x 0.2 & 2000 & 0.8866 (0.8768; 0.8947) & 0.8667 (0.8583; 0.8739) & 0.7859 (0.7719; 0.7992)\\
0.1,0.2; 0.1; 0.2; 0.4 & 4 x 0.2 & 10000 & 0.8977 (0.8885; 0.9048) & 0.8959 (0.8899; 0.8992) & 0.8006 (0.7955; 0.8056)\\
0.1,0.4; 0.1; 0.2; 0.4 & 4 x 0.05 & 200 & 0.5588 (0.5146; 0.6119) & 0.5904 (0.5352; 0.6211) & 0.5505 (0.5000; 0.6036)\\
\addlinespace
0.1,0.4; 0.1; 0.2; 0.4 & 4 x 0.05 & 1000 & 0.6971 (0.6598; 0.7249) & 0.6853 (0.6621; 0.7238) & 0.6690 (0.6491; 0.6985)\\
0.1,0.4; 0.1; 0.2; 0.4 & 4 x 0.05 & 2000 & 0.7180 (0.6913; 0.7446) & 0.7073 (0.6798; 0.7395) & 0.6903 (0.6701; 0.7087)\\
0.1,0.4; 0.1; 0.2; 0.4 & 4 x 0.05 & 10000 & 0.7400 (0.7261; 0.7585) & 0.7332 (0.7082; 0.7617) & 0.7012 (0.6955; 0.7108)\\
0.1,0.4; 0.1; 0.2; 0.4 & 4 x 0.1 & 200 & 0.6622 (0.5951; 0.7451) & 0.6802 (0.6234; 0.7334) & 0.6096 (0.5602; 0.6788)\\
0.1,0.4; 0.1; 0.2; 0.4 & 4 x 0.1 & 1000 & 0.8098 (0.7713; 0.8275) & 0.7965 (0.7463; 0.8326) & 0.7172 (0.6934; 0.7405)\\
\addlinespace
0.1,0.4; 0.1; 0.2; 0.4 & 4 x 0.1 & 2000 & 0.8275 (0.8029; 0.8498) & 0.8344 (0.7741; 0.8490) & 0.7346 (0.7124; 0.7621)\\
0.1,0.4; 0.1; 0.2; 0.4 & 4 x 0.1 & 10000 & 0.8457 (0.8181; 0.8560) & 0.8537 (0.8031; 0.8673) & 0.7537 (0.7407; 0.7677)\\
0.1,0.4; 0.1; 0.2; 0.4 & 4 x 0.2 & 200 & 0.7464 (0.6386; 0.8187) & 0.7430 (0.6756; 0.7844) & 0.6224 (0.5270; 0.6916)\\
0.1,0.4; 0.1; 0.2; 0.4 & 4 x 0.2 & 1000 & 0.8666 (0.8524; 0.8779) & 0.8693 (0.8493; 0.8928) & 0.7602 (0.7327; 0.7815)\\
0.1,0.4; 0.1; 0.2; 0.4 & 4 x 0.2 & 2000 & 0.8827 (0.8711; 0.9003) & 0.8919 (0.8774; 0.9099) & 0.7701 (0.7496; 0.7959)\\
\addlinespace
0.1,0.4; 0.1; 0.2; 0.4 & 4 x 0.2 & 10000 & 0.8934 (0.8811; 0.9086) & 0.9103 (0.9022; 0.9258) & 0.7858 (0.7679; 0.8096)\\
0.2,0.2; 0.1; 0.2; 0.4 & 4 x 0.05 & 200 & 0.5398 (0.5059; 0.6000) & 0.5794 (0.5337; 0.6174) & 0.5448 (0.5000; 0.5902)\\
0.2,0.2; 0.1; 0.2; 0.4 & 4 x 0.05 & 1000 & 0.6767 (0.6478; 0.6935) & 0.6671 (0.6494; 0.6848) & 0.6744 (0.6538; 0.6955)\\
0.2,0.2; 0.1; 0.2; 0.4 & 4 x 0.05 & 2000 & 0.7075 (0.6964; 0.7254) & 0.6718 (0.6637; 0.6891) & 0.6816 (0.6666; 0.6922)\\
0.2,0.2; 0.1; 0.2; 0.4 & 4 x 0.05 & 10000 & 0.7301 (0.7255; 0.7358) & 0.6942 (0.6883; 0.6991) & 0.6923 (0.6855; 0.6977)\\
\addlinespace
0.2,0.2; 0.1; 0.2; 0.4 & 4 x 0.1 & 200 & 0.6093 (0.5340; 0.6759) & 0.6692 (0.6040; 0.7203) & 0.6424 (0.5708; 0.7168)\\
0.2,0.2; 0.1; 0.2; 0.4 & 4 x 0.1 & 1000 & 0.7627 (0.7378; 0.7827) & 0.7317 (0.7120; 0.7504) & 0.7324 (0.7123; 0.7549)\\
0.2,0.2; 0.1; 0.2; 0.4 & 4 x 0.1 & 2000 & 0.7906 (0.7803; 0.8003) & 0.7394 (0.7263; 0.7511) & 0.7440 (0.7317; 0.7537)\\
0.2,0.2; 0.1; 0.2; 0.4 & 4 x 0.1 & 10000 & 0.8022 (0.7978; 0.8050) & 0.7586 (0.7535; 0.7634) & 0.7548 (0.7494; 0.7590)\\
0.2,0.2; 0.1; 0.2; 0.4 & 4 x 0.2 & 200 & 0.6518 (0.5838; 0.7064) & 0.7132 (0.6752; 0.7579) & 0.7122 (0.6432; 0.7566)\\
\addlinespace
0.2,0.2; 0.1; 0.2; 0.4 & 4 x 0.2 & 1000 & 0.8589 (0.8474; 0.8758) & 0.7848 (0.7721; 0.7973) & 0.7889 (0.7722; 0.8031)\\
0.2,0.2; 0.1; 0.2; 0.4 & 4 x 0.2 & 2000 & 0.8687 (0.8615; 0.8775) & 0.7995 (0.7896; 0.8044) & 0.7958 (0.7824; 0.8067)\\
0.2,0.2; 0.1; 0.2; 0.4 & 4 x 0.2 & 10000 & 0.8748 (0.8686; 0.8775) & 0.8324 (0.8247; 0.8379) & 0.8017 (0.7964; 0.8082)\\
0.2,0.4; 0.1; 0.2; 0.4 & 4 x 0.05 & 200 & 0.5485 (0.4959; 0.6187) & 0.5622 (0.5302; 0.6086) & 0.5356 (0.5000; 0.5648)\\
0.2,0.4; 0.1; 0.2; 0.4 & 4 x 0.05 & 1000 & 0.6730 (0.6549; 0.7021) & 0.6718 (0.6504; 0.6970) & 0.6638 (0.6502; 0.6945)\\
\addlinespace
0.2,0.4; 0.1; 0.2; 0.4 & 4 x 0.05 & 2000 & 0.7070 (0.6931; 0.7312) & 0.6932 (0.6767; 0.7062) & 0.6876 (0.6691; 0.7006)\\
0.2,0.4; 0.1; 0.2; 0.4 & 4 x 0.05 & 10000 & 0.7338 (0.7217; 0.7506) & 0.7149 (0.7027; 0.7276) & 0.6998 (0.6924; 0.7067)\\
0.2,0.4; 0.1; 0.2; 0.4 & 4 x 0.1 & 200 & 0.5972 (0.5328; 0.6478) & 0.6428 (0.5925; 0.6905) & 0.6105 (0.5444; 0.6844)\\
0.2,0.4; 0.1; 0.2; 0.4 & 4 x 0.1 & 1000 & 0.7640 (0.7393; 0.7892) & 0.7392 (0.7228; 0.7597) & 0.7288 (0.7099; 0.7493)\\
0.2,0.4; 0.1; 0.2; 0.4 & 4 x 0.1 & 2000 & 0.7906 (0.7729; 0.8092) & 0.7532 (0.7386; 0.7672) & 0.7456 (0.7311; 0.7593)\\
\addlinespace
0.2,0.4; 0.1; 0.2; 0.4 & 4 x 0.1 & 10000 & 0.8038 (0.7944; 0.8172) & 0.7811 (0.7668; 0.7995) & 0.7574 (0.7488; 0.7647)\\
0.2,0.4; 0.1; 0.2; 0.4 & 4 x 0.2 & 200 & 0.6636 (0.6176; 0.7173) & 0.7106 (0.6632; 0.7603) & 0.6853 (0.6446; 0.7266)\\
0.2,0.4; 0.1; 0.2; 0.4 & 4 x 0.2 & 1000 & 0.8552 (0.8346; 0.8729) & 0.8159 (0.7929; 0.8411) & 0.7901 (0.7713; 0.8083)\\
0.2,0.4; 0.1; 0.2; 0.4 & 4 x 0.2 & 2000 & 0.8759 (0.8553; 0.8841) & 0.8342 (0.8244; 0.8535) & 0.7947 (0.7828; 0.8102)\\
0.2,0.4; 0.1; 0.2; 0.4 & 4 x 0.2 & 10000 & 0.8825 (0.8698; 0.8898) & 0.8735 (0.8472; 0.8791) & 0.8121 (0.7990; 0.8188)\\
\addlinespace
0.4,0.4; 0.1; 0.2; 0.4 & 4 x 0.05 & 200 & 0.5534 (0.4935; 0.6192) & 0.5822 (0.5489; 0.6190) & 0.5480 (0.5000; 0.6232)\\
0.4,0.4; 0.1; 0.2; 0.4 & 4 x 0.05 & 1000 & 0.6757 (0.6422; 0.6928) & 0.6630 (0.6479; 0.6785) & 0.6740 (0.6511; 0.6851)\\
0.4,0.4; 0.1; 0.2; 0.4 & 4 x 0.05 & 2000 & 0.7162 (0.7003; 0.7295) & 0.6809 (0.6671; 0.6913) & 0.6920 (0.6694; 0.7010)\\
0.4,0.4; 0.1; 0.2; 0.4 & 4 x 0.05 & 10000 & 0.7320 (0.7250; 0.7364) & 0.6960 (0.6906; 0.7015) & 0.6953 (0.6901; 0.7003)\\
0.4,0.4; 0.1; 0.2; 0.4 & 4 x 0.1 & 200 & 0.6058 (0.5240; 0.6708) & 0.6618 (0.5994; 0.6910) & 0.6451 (0.5624; 0.6961)\\
\addlinespace
0.4,0.4; 0.1; 0.2; 0.4 & 4 x 0.1 & 1000 & 0.7571 (0.7408; 0.7723) & 0.7289 (0.7162; 0.7485) & 0.7422 (0.7211; 0.7544)\\
0.4,0.4; 0.1; 0.2; 0.4 & 4 x 0.1 & 2000 & 0.7931 (0.7813; 0.8027) & 0.7391 (0.7249; 0.7485) & 0.7444 (0.7257; 0.7535)\\
0.4,0.4; 0.1; 0.2; 0.4 & 4 x 0.1 & 10000 & 0.8006 (0.7974; 0.8045) & 0.7555 (0.7509; 0.7606) & 0.7535 (0.7505; 0.7595)\\
0.4,0.4; 0.1; 0.2; 0.4 & 4 x 0.2 & 200 & 0.6672 (0.6100; 0.7296) & 0.7258 (0.6947; 0.7662) & 0.7384 (0.6855; 0.7706)\\
0.4,0.4; 0.1; 0.2; 0.4 & 4 x 0.2 & 1000 & 0.8550 (0.8377; 0.8672) & 0.7916 (0.7710; 0.8076) & 0.7904 (0.7740; 0.8081)\\
\addlinespace
0.4,0.4; 0.1; 0.2; 0.4 & 4 x 0.2 & 2000 & 0.8652 (0.8586; 0.8720) & 0.7954 (0.7862; 0.8037) & 0.7962 (0.7791; 0.8069)\\
0.4,0.4; 0.1; 0.2; 0.4 & 4 x 0.2 & 10000 & 0.8724 (0.8687; 0.8768) & 0.8176 (0.8134; 0.8238) & 0.8061 (0.8016; 0.8114)\\*
\end{longtable}
\end{ThreePartTable}

\FloatBarrier

\begin{ThreePartTable}
\begin{TableNotes}
\item \textit{Note: } 
\item Performance of the algorithms \texttt{MBMDRC}, \texttt{RANGER}, and \texttt{GLMNET} measured as AUC over 50 replicates in scenario 7. The median of the AUC and the 25
\end{TableNotes}
\begin{longtable}{llrlll}
\caption{\label{tab:performance-table-scenario5}Performance in scenario 5.}\\
\toprule
MAF & $h^2$ & $n$ & MBMDRC & RANGER & GLMNET\\
\midrule
\endfirsthead
\caption[]{\label{tab:performance-table-scenario5}Performance in scenario 5. \textit{(continued)}}\\
\toprule
MAF & $h^2$ & $n$ & MBMDRC & RANGER & GLMNET\\
\midrule
\endhead
\
\endfoot
\bottomrule
\insertTableNotes
\endlastfoot
0.1,0.1; 0.1,0.1 & 2 x 0.05 & 200 & 0.5018 (0.4482; 0.5447) & 0.4742 (0.4370; 0.5245) & 0.5000 (0.4776; 0.5076)\\
0.1,0.1; 0.1,0.1 & 2 x 0.05 & 1000 & 0.5729 (0.5381; 0.6217) & 0.5078 (0.4883; 0.5334) & 0.5000 (0.4995; 0.5050)\\
0.1,0.1; 0.1,0.1 & 2 x 0.05 & 2000 & 0.6341 (0.6177; 0.6443) & 0.5091 (0.4975; 0.5215) & 0.5000 (0.4963; 0.5000)\\
0.1,0.1; 0.1,0.1 & 2 x 0.05 & 10000 & 0.6423 (0.6365; 0.6460) & 0.5369 (0.5291; 0.5487) & 0.5000 (0.4992; 0.5000)\\
0.1,0.1; 0.1,0.1 & 2 x 0.1 & 200 & 0.5295 (0.4705; 0.5998) & 0.5036 (0.4469; 0.5540) & 0.5000 (0.4847; 0.5120)\\
\addlinespace
0.1,0.1; 0.1,0.1 & 2 x 0.1 & 1000 & 0.7614 (0.7394; 0.7796) & 0.6542 (0.6093; 0.6872) & 0.5380 (0.5000; 0.5683)\\
0.1,0.1; 0.1,0.1 & 2 x 0.1 & 2000 & 0.7672 (0.7431; 0.7866) & 0.7312 (0.6911; 0.7515) & 0.5461 (0.5123; 0.6044)\\
0.1,0.1; 0.1,0.1 & 2 x 0.1 & 10000 & 0.7690 (0.7514; 0.7873) & 0.7614 (0.7486; 0.7851) & 0.6103 (0.5457; 0.6428)\\
0.1,0.1; 0.1,0.1 & 2 x 0.2 & 200 & 0.6838 (0.5958; 0.7467) & 0.5660 (0.5059; 0.6247) & 0.5298 (0.5000; 0.5716)\\
0.1,0.1; 0.1,0.1 & 2 x 0.2 & 1000 & 0.8552 (0.8317; 0.8703) & 0.8452 (0.8145; 0.8637) & 0.6320 (0.6023; 0.6588)\\
\addlinespace
0.1,0.1; 0.1,0.1 & 2 x 0.2 & 2000 & 0.8508 (0.8330; 0.8812) & 0.8499 (0.8258; 0.8774) & 0.6519 (0.6124; 0.6992)\\
0.1,0.1; 0.1,0.1 & 2 x 0.2 & 10000 & 0.8481 (0.8390; 0.8796) & 0.8461 (0.8358; 0.8786) & 0.6942 (0.6252; 0.7088)\\
0.1,0.1; 0.2,0.2 & 2 x 0.05 & 200 & 0.4951 (0.4496; 0.5640) & 0.4678 (0.4320; 0.5213) & 0.5000 (0.4804; 0.5000)\\
0.1,0.1; 0.2,0.2 & 2 x 0.05 & 1000 & 0.5874 (0.5502; 0.6151) & 0.5051 (0.4891; 0.5286) & 0.5000 (0.4985; 0.5088)\\
0.1,0.1; 0.2,0.2 & 2 x 0.05 & 2000 & 0.6502 (0.6366; 0.6650) & 0.5089 (0.4969; 0.5203) & 0.5000 (0.4969; 0.5000)\\
\addlinespace
0.1,0.1; 0.2,0.2 & 2 x 0.05 & 10000 & 0.6575 (0.6531; 0.6615) & 0.5362 (0.5284; 0.5494) & 0.5000 (0.4972; 0.5000)\\
0.1,0.1; 0.2,0.2 & 2 x 0.1 & 200 & 0.5434 (0.4746; 0.6322) & 0.4806 (0.4463; 0.5181) & 0.5000 (0.4820; 0.5094)\\
0.1,0.1; 0.2,0.2 & 2 x 0.1 & 1000 & 0.7782 (0.7485; 0.7954) & 0.5772 (0.5486; 0.6289) & 0.5055 (0.5000; 0.5335)\\
0.1,0.1; 0.2,0.2 & 2 x 0.1 & 2000 & 0.7826 (0.7661; 0.7942) & 0.6532 (0.5906; 0.6972) & 0.5084 (0.5000; 0.5411)\\
0.1,0.1; 0.2,0.2 & 2 x 0.1 & 10000 & 0.7850 (0.7674; 0.7937) & 0.7546 (0.7319; 0.7666) & 0.5387 (0.5230; 0.5926)\\
\addlinespace
0.1,0.1; 0.2,0.2 & 2 x 0.2 & 200 & 0.6704 (0.5762; 0.7401) & 0.4882 (0.4579; 0.5323) & 0.5000 (0.4676; 0.5189)\\
0.1,0.1; 0.2,0.2 & 2 x 0.2 & 1000 & 0.8537 (0.8424; 0.8643) & 0.7350 (0.6601; 0.7630) & 0.5339 (0.5000; 0.5626)\\
0.1,0.1; 0.2,0.2 & 2 x 0.2 & 2000 & 0.8550 (0.8463; 0.8652) & 0.7872 (0.7663; 0.8166) & 0.5426 (0.5159; 0.6173)\\
0.1,0.1; 0.2,0.2 & 2 x 0.2 & 10000 & 0.8585 (0.8501; 0.8642) & 0.8558 (0.8455; 0.8617) & 0.5507 (0.5379; 0.6327)\\
0.1,0.1; 0.4,0.4 & 2 x 0.05 & 200 & 0.4884 (0.4392; 0.5440) & 0.4734 (0.4356; 0.5165) & 0.5000 (0.4818; 0.5000)\\
\addlinespace
0.1,0.1; 0.4,0.4 & 2 x 0.05 & 1000 & 0.5746 (0.5330; 0.6174) & 0.5058 (0.4876; 0.5304) & 0.5000 (0.4980; 0.5088)\\
0.1,0.1; 0.4,0.4 & 2 x 0.05 & 2000 & 0.6508 (0.6396; 0.6662) & 0.5054 (0.4956; 0.5175) & 0.5000 (0.4986; 0.5019)\\
0.1,0.1; 0.4,0.4 & 2 x 0.05 & 10000 & 0.6586 (0.6564; 0.6636) & 0.5381 (0.5267; 0.5448) & 0.5000 (0.4990; 0.5000)\\
0.1,0.1; 0.4,0.4 & 2 x 0.1 & 200 & 0.5120 (0.4561; 0.5588) & 0.4814 (0.4441; 0.5067) & 0.5000 (0.4747; 0.5022)\\
0.1,0.1; 0.4,0.4 & 2 x 0.1 & 1000 & 0.7421 (0.7240; 0.7644) & 0.5350 (0.5114; 0.5658) & 0.5000 (0.4966; 0.5027)\\
\addlinespace
0.1,0.1; 0.4,0.4 & 2 x 0.1 & 2000 & 0.7511 (0.7395; 0.7616) & 0.5651 (0.5417; 0.6097) & 0.5000 (0.5000; 0.5145)\\
0.1,0.1; 0.4,0.4 & 2 x 0.1 & 10000 & 0.7575 (0.7499; 0.7658) & 0.6944 (0.6776; 0.7229) & 0.5129 (0.5000; 0.5278)\\
0.1,0.1; 0.4,0.4 & 2 x 0.2 & 200 & 0.6307 (0.5475; 0.7046) & 0.4970 (0.4664; 0.5456) & 0.5000 (0.4991; 0.5000)\\
0.1,0.1; 0.4,0.4 & 2 x 0.2 & 1000 & 0.8476 (0.8349; 0.8604) & 0.6328 (0.5583; 0.7149) & 0.5000 (0.5000; 0.5367)\\
0.1,0.1; 0.4,0.4 & 2 x 0.2 & 2000 & 0.8460 (0.8354; 0.8572) & 0.6940 (0.6220; 0.7598) & 0.5113 (0.5000; 0.5448)\\
\addlinespace
0.1,0.1; 0.4,0.4 & 2 x 0.2 & 10000 & 0.8514 (0.8470; 0.8557) & 0.8375 (0.8132; 0.8467) & 0.5466 (0.5289; 0.5896)\\
0.2,0.2; 0.2,0.2 & 2 x 0.05 & 200 & 0.5099 (0.4680; 0.5507) & 0.4626 (0.4282; 0.5151) & 0.5000 (0.4733; 0.5005)\\
0.2,0.2; 0.2,0.2 & 2 x 0.05 & 1000 & 0.5775 (0.5389; 0.6184) & 0.5085 (0.4900; 0.5290) & 0.5000 (0.4986; 0.5109)\\
0.2,0.2; 0.2,0.2 & 2 x 0.05 & 2000 & 0.6552 (0.6359; 0.6669) & 0.5055 (0.4942; 0.5143) & 0.5000 (0.4972; 0.5000)\\
0.2,0.2; 0.2,0.2 & 2 x 0.05 & 10000 & 0.6631 (0.6592; 0.6696) & 0.5383 (0.5255; 0.5552) & 0.5000 (0.4978; 0.5000)\\
\addlinespace
0.2,0.2; 0.2,0.2 & 2 x 0.1 & 200 & 0.5104 (0.4559; 0.5708) & 0.4614 (0.4290; 0.5134) & 0.5000 (0.4684; 0.5000)\\
0.2,0.2; 0.2,0.2 & 2 x 0.1 & 1000 & 0.7166 (0.6938; 0.7337) & 0.5152 (0.4986; 0.5377) & 0.5000 (0.5000; 0.5099)\\
0.2,0.2; 0.2,0.2 & 2 x 0.1 & 2000 & 0.7253 (0.7126; 0.7389) & 0.5244 (0.5094; 0.5381) & 0.5000 (0.4959; 0.5021)\\
0.2,0.2; 0.2,0.2 & 2 x 0.1 & 10000 & 0.7308 (0.7238; 0.7353) & 0.5804 (0.5575; 0.5975) & 0.5000 (0.4988; 0.5000)\\
0.2,0.2; 0.2,0.2 & 2 x 0.2 & 200 & 0.5273 (0.4582; 0.6190) & 0.4706 (0.4218; 0.5199) & 0.5000 (0.4745; 0.5000)\\
\addlinespace
0.2,0.2; 0.2,0.2 & 2 x 0.2 & 1000 & 0.7994 (0.7873; 0.8181) & 0.5352 (0.5108; 0.5650) & 0.5000 (0.4970; 0.5020)\\
0.2,0.2; 0.2,0.2 & 2 x 0.2 & 2000 & 0.8103 (0.7986; 0.8203) & 0.5497 (0.5291; 0.6064) & 0.5000 (0.4929; 0.5000)\\
0.2,0.2; 0.2,0.2 & 2 x 0.2 & 10000 & 0.8139 (0.8098; 0.8193) & 0.6950 (0.6361; 0.7439) & 0.5000 (0.5000; 0.5013)\\
0.2,0.2; 0.4,0.4 & 2 x 0.05 & 200 & 0.4968 (0.4626; 0.5564) & 0.4714 (0.4342; 0.5195) & 0.5000 (0.4948; 0.5072)\\
0.2,0.2; 0.4,0.4 & 2 x 0.05 & 1000 & 0.5698 (0.5280; 0.6089) & 0.5057 (0.4867; 0.5273) & 0.5000 (0.5000; 0.5108)\\
\addlinespace
0.2,0.2; 0.4,0.4 & 2 x 0.05 & 2000 & 0.6567 (0.6398; 0.6695) & 0.5030 (0.4913; 0.5169) & 0.5000 (0.4970; 0.5000)\\
0.2,0.2; 0.4,0.4 & 2 x 0.05 & 10000 & 0.6683 (0.6625; 0.6728) & 0.5312 (0.5240; 0.5405) & 0.5000 (0.4977; 0.5000)\\
0.2,0.2; 0.4,0.4 & 2 x 0.1 & 200 & 0.5064 (0.4639; 0.5667) & 0.4700 (0.4317; 0.5068) & 0.5000 (0.4735; 0.5000)\\
0.2,0.2; 0.4,0.4 & 2 x 0.1 & 1000 & 0.7116 (0.6958; 0.7333) & 0.5149 (0.4927; 0.5345) & 0.5000 (0.5000; 0.5086)\\
0.2,0.2; 0.4,0.4 & 2 x 0.1 & 2000 & 0.7294 (0.7194; 0.7409) & 0.5172 (0.5066; 0.5295) & 0.5000 (0.4951; 0.5022)\\
\addlinespace
0.2,0.2; 0.4,0.4 & 2 x 0.1 & 10000 & 0.7352 (0.7308; 0.7401) & 0.5718 (0.5553; 0.5941) & 0.5000 (0.4975; 0.5003)\\
0.2,0.2; 0.4,0.4 & 2 x 0.2 & 200 & 0.5403 (0.4834; 0.6339) & 0.4738 (0.4309; 0.5153) & 0.5000 (0.4612; 0.5000)\\
0.2,0.2; 0.4,0.4 & 2 x 0.2 & 1000 & 0.8005 (0.7917; 0.8084) & 0.5323 (0.5003; 0.5569) & 0.5000 (0.4949; 0.5036)\\
0.2,0.2; 0.4,0.4 & 2 x 0.2 & 2000 & 0.8164 (0.8048; 0.8255) & 0.5517 (0.5364; 0.5741) & 0.5000 (0.4963; 0.5024)\\
0.2,0.2; 0.4,0.4 & 2 x 0.2 & 10000 & 0.8174 (0.8140; 0.8213) & 0.6909 (0.6706; 0.7253) & 0.5000 (0.4989; 0.5028)\\
\addlinespace
0.4,0.4; 0.4,0.4 & 2 x 0.05 & 200 & 0.4941 (0.4540; 0.5431) & 0.4722 (0.4321; 0.5202) & 0.5000 (0.4787; 0.5051)\\
0.4,0.4; 0.4,0.4 & 2 x 0.05 & 1000 & 0.5850 (0.5344; 0.6367) & 0.5077 (0.4889; 0.5258) & 0.5000 (0.4963; 0.5075)\\
0.4,0.4; 0.4,0.4 & 2 x 0.05 & 2000 & 0.6585 (0.6345; 0.6714) & 0.5061 (0.4963; 0.5147) & 0.5000 (0.4959; 0.5005)\\
0.4,0.4; 0.4,0.4 & 2 x 0.05 & 10000 & 0.6675 (0.6629; 0.6719) & 0.5341 (0.5253; 0.5438) & 0.5000 (0.4989; 0.5000)\\
0.4,0.4; 0.4,0.4 & 2 x 0.1 & 200 & 0.5147 (0.4555; 0.5712) & 0.4624 (0.4277; 0.5220) & 0.5000 (0.4857; 0.5000)\\
\addlinespace
0.4,0.4; 0.4,0.4 & 2 x 0.1 & 1000 & 0.7054 (0.6815; 0.7273) & 0.5104 (0.4918; 0.5328) & 0.5000 (0.4993; 0.5065)\\
0.4,0.4; 0.4,0.4 & 2 x 0.1 & 2000 & 0.7302 (0.7191; 0.7403) & 0.5157 (0.5024; 0.5292) & 0.5000 (0.4949; 0.5007)\\
0.4,0.4; 0.4,0.4 & 2 x 0.1 & 10000 & 0.7332 (0.7290; 0.7391) & 0.5750 (0.5604; 0.6000) & 0.5000 (0.5000; 0.5008)\\
0.4,0.4; 0.4,0.4 & 2 x 0.2 & 200 & 0.5200 (0.4829; 0.6178) & 0.4764 (0.4308; 0.5242) & 0.5000 (0.4765; 0.5000)\\
0.4,0.4; 0.4,0.4 & 2 x 0.2 & 1000 & 0.8051 (0.7879; 0.8191) & 0.5261 (0.5002; 0.5464) & 0.5000 (0.4962; 0.5042)\\
\addlinespace
0.4,0.4; 0.4,0.4 & 2 x 0.2 & 2000 & 0.8125 (0.8035; 0.8229) & 0.5431 (0.5315; 0.5696) & 0.5000 (0.4959; 0.5005)\\
0.4,0.4; 0.4,0.4 & 2 x 0.2 & 10000 & 0.8151 (0.8111; 0.8184) & 0.6761 (0.6345; 0.7042) & 0.5000 (0.4999; 0.5004)\\*
\end{longtable}
\end{ThreePartTable}

\FloatBarrier
\begin{table}[t]

\caption{\label{tab:performance-table-scenario6}Performance in scenario 6.}
\centering
\resizebox{\linewidth}{!}{
\begin{threeparttable}
\begin{tabular}{llrlll}
\toprule
MAF & $h^2$ & $n$ & MBMDRC & RANGER & GLMNET\\
\midrule
0.1,0.1; 0.2,0.2; 0.4,0.4 & 3 x 0.05 & 200 & 0.5034 (0.4565; 0.5560) & 0.4964 (0.4546; 0.5219) & 0.5000 (0.4755; 0.5000)\\
0.1,0.1; 0.2,0.2; 0.4,0.4 & 3 x 0.05 & 1000 & 0.5897 (0.5347; 0.6276) & 0.5110 (0.4895; 0.5291) & 0.5000 (0.4955; 0.5074)\\
0.1,0.1; 0.2,0.2; 0.4,0.4 & 3 x 0.05 & 2000 & 0.6772 (0.6619; 0.6919) & 0.5106 (0.4941; 0.5295) & 0.5000 (0.4940; 0.5019)\\
0.1,0.1; 0.2,0.2; 0.4,0.4 & 3 x 0.05 & 10000 & 0.7013 (0.6955; 0.7068) & 0.5461 (0.5362; 0.5602) & 0.5000 (0.5000; 0.5015)\\
0.1,0.1; 0.2,0.2; 0.4,0.4 & 3 x 0.1 & 200 & 0.5425 (0.4950; 0.5902) & 0.5102 (0.4584; 0.5539) & 0.5000 (0.4824; 0.5178)\\
\addlinespace
0.1,0.1; 0.2,0.2; 0.4,0.4 & 3 x 0.1 & 1000 & 0.7911 (0.7742; 0.8189) & 0.5843 (0.5448; 0.6251) & 0.5164 (0.5000; 0.5364)\\
0.1,0.1; 0.2,0.2; 0.4,0.4 & 3 x 0.1 & 2000 & 0.8061 (0.7912; 0.8219) & 0.6335 (0.5889; 0.6764) & 0.5078 (0.5000; 0.5315)\\
0.1,0.1; 0.2,0.2; 0.4,0.4 & 3 x 0.1 & 10000 & 0.8164 (0.8050; 0.8282) & 0.7408 (0.7048; 0.7653) & 0.5543 (0.5237; 0.5867)\\
0.1,0.1; 0.2,0.2; 0.4,0.4 & 3 x 0.2 & 200 & 0.6134 (0.5345; 0.7101) & 0.5022 (0.4708; 0.5483) & 0.5000 (0.4954; 0.5136)\\
0.1,0.1; 0.2,0.2; 0.4,0.4 & 3 x 0.2 & 1000 & 0.8785 (0.8666; 0.8862) & 0.6639 (0.5814; 0.7149) & 0.5163 (0.5000; 0.5657)\\
\addlinespace
0.1,0.1; 0.2,0.2; 0.4,0.4 & 3 x 0.2 & 2000 & 0.8834 (0.8714; 0.8893) & 0.7453 (0.6979; 0.7692) & 0.5381 (0.5001; 0.6176)\\
0.1,0.1; 0.2,0.2; 0.4,0.4 & 3 x 0.2 & 10000 & 0.8902 (0.8836; 0.8949) & 0.8408 (0.8293; 0.8576) & 0.5590 (0.5324; 0.6303)\\
\bottomrule
\end{tabular}
\begin{tablenotes}
\item \textit{Note: } 
\item Performance of the algorithms \texttt{MBMDRC}, \texttt{RANGER}, and \texttt{GLMNET} measured as AUC over 50 replicates in scenario 7. The median of the AUC and the 25
\end{tablenotes}
\end{threeparttable}}
\end{table}

\FloatBarrier
\begin{table}[t]

\caption{\label{tab:performance-table-scenario7}Performance in scenario 7.}
\centering
\resizebox{\linewidth}{!}{
\begin{threeparttable}
\begin{tabular}{llrlll}
\toprule
MAF & $h^2$ & $n$ & MBMDRC & RANGER & GLMNET\\
\midrule
0.1,0.1; 0.2,0.2; 0.4,0.4; 0.1; 0.2; 0.4 & 6 x 0.05 & 200 & 0.5398 (0.4819; 0.5874) & 0.5316 (0.5021; 0.5890) & 0.5282 (0.5000; 0.5956)\\
0.1,0.1; 0.2,0.2; 0.4,0.4; 0.1; 0.2; 0.4 & 6 x 0.05 & 1000 & 0.6594 (0.6263; 0.6825) & 0.6523 (0.6358; 0.6710) & 0.6596 (0.6374; 0.6772)\\
0.1,0.1; 0.2,0.2; 0.4,0.4; 0.1; 0.2; 0.4 & 6 x 0.05 & 2000 & 0.7407 (0.7150; 0.7602) & 0.6688 (0.6521; 0.6817) & 0.6687 (0.6511; 0.6821)\\
0.1,0.1; 0.2,0.2; 0.4,0.4; 0.1; 0.2; 0.4 & 6 x 0.05 & 10000 & 0.7707 (0.7650; 0.7781) & 0.6938 (0.6882; 0.7093) & 0.6811 (0.6704; 0.6871)\\
0.1,0.1; 0.2,0.2; 0.4,0.4; 0.1; 0.2; 0.4 & 6 x 0.1 & 200 & 0.6009 (0.5340; 0.6414) & 0.6114 (0.5833; 0.6748) & 0.6176 (0.5736; 0.6615)\\
\addlinespace
0.1,0.1; 0.2,0.2; 0.4,0.4; 0.1; 0.2; 0.4 & 6 x 0.1 & 1000 & 0.8087 (0.7657; 0.8445) & 0.7167 (0.6968; 0.7320) & 0.7109 (0.6912; 0.7243)\\
0.1,0.1; 0.2,0.2; 0.4,0.4; 0.1; 0.2; 0.4 & 6 x 0.1 & 2000 & 0.8533 (0.8385; 0.8697) & 0.7326 (0.7219; 0.7475) & 0.7213 (0.7078; 0.7441)\\
0.1,0.1; 0.2,0.2; 0.4,0.4; 0.1; 0.2; 0.4 & 6 x 0.1 & 10000 & 0.8639 (0.8554; 0.8737) & 0.7732 (0.7631; 0.7838) & 0.7360 (0.7283; 0.7465)\\
0.1,0.1; 0.2,0.2; 0.4,0.4; 0.1; 0.2; 0.4 & 6 x 0.2 & 200 & 0.6080 (0.5592; 0.6636) & 0.6376 (0.5958; 0.7031) & 0.6282 (0.5608; 0.6964)\\
0.1,0.1; 0.2,0.2; 0.4,0.4; 0.1; 0.2; 0.4 & 6 x 0.2 & 1000 & 0.8861 (0.8577; 0.9070) & 0.7510 (0.7328; 0.7710) & 0.7374 (0.7208; 0.7526)\\
\addlinespace
0.1,0.1; 0.2,0.2; 0.4,0.4; 0.1; 0.2; 0.4 & 6 x 0.2 & 2000 & 0.9101 (0.9025; 0.9197) & 0.7858 (0.7706; 0.7996) & 0.7491 (0.7332; 0.7616)\\
0.1,0.1; 0.2,0.2; 0.4,0.4; 0.1; 0.2; 0.4 & 6 x 0.2 & 10000 & 0.9231 (0.9165; 0.9256) & 0.8387 (0.8323; 0.8457) & 0.7608 (0.7522; 0.7651)\\
\bottomrule
\end{tabular}
\begin{tablenotes}
\item \textit{Note: } 
\item Performance of the algorithms \texttt{MBMDRC}, \texttt{RANGER}, and \texttt{GLMNET} measured as AUC over 50 replicates in scenario 7. The median of the AUC and the 25
\end{tablenotes}
\end{threeparttable}}
\end{table}

\FloatBarrier
\begin{table}[t]

\caption{\label{tab:performance-table-scenario8}Performance in scenario 8.}
\centering
\resizebox{\linewidth}{!}{
\begin{threeparttable}
\begin{tabular}{llrlll}
\toprule
MAF & $h^2$ & $n$ & MBMDRC & RANGER & GLMNET\\
\midrule
0.1,0.2,0.4; 0.1; 0.2; 0.4 & 4 x 0.05 & 200 & 0.5404 (0.4890; 0.5999) & 0.5536 (0.5202; 0.5940) & 0.5134 (0.5000; 0.5910)\\
0.1,0.2,0.4; 0.1; 0.2; 0.4 & 4 x 0.05 & 1000 & 0.6540 (0.6300; 0.6796) & 0.6628 (0.6401; 0.6753) & 0.6604 (0.6409; 0.6761)\\
0.1,0.2,0.4; 0.1; 0.2; 0.4 & 4 x 0.05 & 2000 & 0.6939 (0.6790; 0.7106) & 0.6853 (0.6687; 0.6922) & 0.6783 (0.6669; 0.6915)\\
0.1,0.2,0.4; 0.1; 0.2; 0.4 & 4 x 0.05 & 10000 & 0.7198 (0.7128; 0.7280) & 0.7227 (0.6999; 0.7345) & 0.6964 (0.6872; 0.7013)\\
0.1,0.2,0.4; 0.1; 0.2; 0.4 & 4 x 0.1 & 200 & 0.5800 (0.5192; 0.6256) & 0.6066 (0.5653; 0.6639) & 0.5950 (0.5533; 0.6309)\\
\addlinespace
0.1,0.2,0.4; 0.1; 0.2; 0.4 & 4 x 0.1 & 1000 & 0.7212 (0.6997; 0.7402) & 0.7108 (0.6989; 0.7260) & 0.7074 (0.6916; 0.7291)\\
0.1,0.2,0.4; 0.1; 0.2; 0.4 & 4 x 0.1 & 2000 & 0.7605 (0.7456; 0.7755) & 0.7249 (0.7142; 0.7393) & 0.7200 (0.7057; 0.7356)\\
0.1,0.2,0.4; 0.1; 0.2; 0.4 & 4 x 0.1 & 10000 & 0.7737 (0.7666; 0.7812) & 0.7731 (0.7599; 0.7829) & 0.7371 (0.7307; 0.7416)\\
\bottomrule
\end{tabular}
\begin{tablenotes}
\item \textit{Note: } 
\item Performance of the algorithms \texttt{MBMDRC}, \texttt{RANGER}, and \texttt{GLMNET} measured as AUC over 50 replicates in scenario 7. The median of the AUC and the 25
\end{tablenotes}
\end{threeparttable}}
\end{table}

\FloatBarrier

\hypertarget{references}{%
\section*{References}\label{references}}
\addcontentsline{toc}{section}{References}

\hypertarget{refs}{}
\leavevmode\hypertarget{ref-Konig2017}{}%
1. König IR, Fuchs O, Hansen G, Mutius E von, Kopp MV. What is precision medicine? Eur Respir J. 2017; doi:\href{https://doi.org/10.1183/13993003.00391-2017}{10.1183/13993003.00391-2017}.

\leavevmode\hypertarget{ref-Redekop2013}{}%
2. Redekop WK, Mladsi D. The Faces of Personalized Medicine: A Framework for Understanding Its Meaning and Scope. Value Heal. 2013; doi:\href{https://doi.org/10.1016/j.jval.2013.06.005}{10.1016/j.jval.2013.06.005}.

\leavevmode\hypertarget{ref-Burke2014}{}%
3. Burke W, Brown Trinidad S, Press NA. Essential elements of personalized medicine. Urol Oncol Semin Orig Investig. 2014; doi:\href{https://doi.org/10.1016/j.urolonc.2013.09.002}{10.1016/j.urolonc.2013.09.002}.

\leavevmode\hypertarget{ref-Chen2015}{}%
4. Chen C, He M, Zhu Y, Shi L, Wang X. Five critical elements to ensure the precision medicine. Cancer Metastasis Rev. 2015; doi:\href{https://doi.org/10.1007/s10555-015-9555-3}{10.1007/s10555-015-9555-3}.

\leavevmode\hypertarget{ref-Seymour2017}{}%
5. Seymour CW, Gomez H, Chang C-CH, Clermont G, Kellum JA, Kennedy J, et al. Precision medicine for all? Challenges and opportunities for a precision medicine approach to critical illness. Crit Care. 2017; doi:\href{https://doi.org/10.1186/s13054-017-1836-5}{10.1186/s13054-017-1836-5}.

\leavevmode\hypertarget{ref-Lin2017}{}%
6. Lin J-Z, Long J-Y, Wang A-Q, Zheng Y, Zhao H-T. Precision medicine: In need of guidance and surveillance. World J Gastroenterol. 2017; doi:\href{https://doi.org/10.3748/wjg.v23.i28.5045}{10.3748/wjg.v23.i28.5045}.

\leavevmode\hypertarget{ref-Jordan2018}{}%
7. Jordan DM, Do R. Using Full Genomic Information to Predict Disease: Breaking Down the Barriers Between Complex and Mendelian Diseases. Annu Rev Genomics Hum Genet. 2018; doi:\href{https://doi.org/10.1146/annurev-genom-083117-021136}{10.1146/annurev-genom-083117-021136}.

\leavevmode\hypertarget{ref-Cordell2002}{}%
8. Cordell HJ. Epistasis: what it means, what it doesn't mean, and statistical methods to detect it in humans. Hum Mol Genet. 2002; doi:\href{https://doi.org/10.1093/hmg/11.20.2463}{10.1093/hmg/11.20.2463}.

\leavevmode\hypertarget{ref-Cordell2009}{}%
9. Cordell HJ. Detecting gene-gene interactions that underlie human diseases. Nat Rev Genet. 2009; doi:\href{https://doi.org/10.1038/nrg2579}{10.1038/nrg2579}.

\leavevmode\hypertarget{ref-Thomas2010}{}%
10. Thomas D. Gene--environment-wide association studies: emerging approaches. Nat Rev Genet. 2010; doi:\href{https://doi.org/10.1038/nrg2764}{10.1038/nrg2764}.

\leavevmode\hypertarget{ref-Ritchie2018}{}%
11. Ritchie MD, Van Steen K. The search for gene-gene interactions in genome-wide association studies: challenges in abundance of methods, practical considerations, and biological interpretation. Ann Transl Med. 2018; doi:\href{https://doi.org/10.21037/atm.2018.04.05}{10.21037/atm.2018.04.05}.

\leavevmode\hypertarget{ref-Eichler2010}{}%
12. Eichler EE, Flint J, Gibson G, Kong A, Leal SM, Moore JH, et al. Missing heritability and strategies for finding the underlying causes of complex disease. Nat Rev Genet. 2010; doi:\href{https://doi.org/10.1038/nrg2809}{10.1038/nrg2809}.

\leavevmode\hypertarget{ref-Zschiedrich2009}{}%
13. Zschiedrich K, König IR, Brüggemann N, Kock N, Kasten M, Leenders KL, et al. MDR1 variants and risk of Parkinson disease. J Neurol. 2009; doi:\href{https://doi.org/10.1007/s00415-009-0089-x}{10.1007/s00415-009-0089-x}.

\leavevmode\hypertarget{ref-Aschard2016}{}%
14. Aschard H. A perspective on interaction effects in genetic association studies. Genet Epidemiol. 2016; doi:\href{https://doi.org/10.1002/gepi.21989}{10.1002/gepi.21989}.

\leavevmode\hypertarget{ref-Hoerl1970}{}%
15. Hoerl AE, Kennard RW. Ridge Regression: Biased Estimation for Nonorthogonal Problems. Technometrics. 1970; doi:\href{https://doi.org/10.1080/00401706.1970.10488634}{10.1080/00401706.1970.10488634}.

\leavevmode\hypertarget{ref-Tibshirani1996}{}%
16. Tibshirani R. Regression Selection and Shrinkage via the Lasso. J R Stat Soc B. 1996;58:267--88.

\leavevmode\hypertarget{ref-Breiman2001}{}%
17. Breiman L. Random Forests. Mach Learn. 2001; doi:\href{https://doi.org/10.1023/A:1010933404324}{10.1023/A:1010933404324}.

\leavevmode\hypertarget{ref-Ishwaran2015}{}%
18. Ishwaran H. The effect of splitting on random forests. Mach Learn. 2015; doi:\href{https://doi.org/10.1007/s10994-014-5451-2}{10.1007/s10994-014-5451-2}.

\leavevmode\hypertarget{ref-Wright2016}{}%
19. Wright MN, Ziegler A, König IR. Do little interactions get lost in dark random forests? BMC Bioinformatics. 2016; doi:\href{https://doi.org/10.1186/s12859-016-0995-8}{10.1186/s12859-016-0995-8}.

\leavevmode\hypertarget{ref-Friedman2010}{}%
20. Friedman J, Hastie T, Tibshirani R. Regularization Paths for Generalized Linear Models via Coordinate Descent. J Stat Softw. 2010; doi:\href{https://doi.org/10.18637/jss.v033.i01}{10.18637/jss.v033.i01}.

\leavevmode\hypertarget{ref-Koo2015}{}%
21. Koo CL, Liew MJ, Mohamad MS, Salleh AHM, Deris S, Ibrahim Z, et al. Software for detecting gene-gene interactions in genome wide association studies. Biotechnol Bioprocess Eng. 2015; doi:\href{https://doi.org/10.1007/s12257-015-0064-6}{10.1007/s12257-015-0064-6}.

\leavevmode\hypertarget{ref-Ritchie2001}{}%
22. Ritchie MD, Hahn LW, Roodi N, Bailey LR, Dupont W., Parl FF, et al. Multifactor-Dimensionality Reduction Reveals High-Order Interactions among Estrogen-Metabolism Genes in Sporadic Breast Cancer. Am J Hum Genet. 2001; doi:\href{https://doi.org/10.1086/321276}{10.1086/321276}.

\leavevmode\hypertarget{ref-Gola2015}{}%
23. Gola D, Mahachie John JM, Steen K van, König IR. A roadmap to multifactor dimensionality reduction methods. Brief Bioinform. 2016; doi:\href{https://doi.org/10.1093/bib/bbv038}{10.1093/bib/bbv038}.

\leavevmode\hypertarget{ref-Calle2008a}{}%
24. Calle ML, Urrea Gales V, Malats i Riera N, Van Steen K. MB-MDR: Model-Based Multifactor Dimensionality Reduction for detecting interactions in high-dimensional genomic data. 2008. \url{http://dspace.uvic.cat/handle/10854/408}. Accessed 1 Nov 2018.

\leavevmode\hypertarget{ref-Urbanowicz2012b}{}%
25. Urbanowicz RJ, Kiralis J, Sinnott-Armstrong N a, Heberling T, Fisher JM, Moore JH. GAMETES: a fast, direct algorithm for generating pure, strict, epistatic models with random architectures. BioData Min. 2012; doi:\href{https://doi.org/10.1186/1756-0381-5-16}{10.1186/1756-0381-5-16}.

\leavevmode\hypertarget{ref-Kraft2007}{}%
26. Kraft P. Statistical issues in epidemiological studies of gene-environment interaction. 2007. \url{https://hstalks.com/bs/79/}. Accessed 1 Nov 2018.

\leavevmode\hypertarget{ref-Gola347096}{}%
27. Gola D, Hessler N, Schwaninger M, Ziegler A, König IR. Evaluating predictive biomarkers for a binary outcome with linear versus logistic regression - Practical recommendations for the choice of the model. bioRxiv. 2018; doi:\href{https://doi.org/10.1101/347096}{10.1101/347096}.

\leavevmode\hypertarget{ref-Calle2008}{}%
28. Calle ML, Urrea V, Vellalta G, Malats N, Steen KV. Improving strategies for detecting genetic patterns of disease susceptibility in association studies. Stat Med. 2008; doi:\href{https://doi.org/10.1002/sim.3431}{10.1002/sim.3431}.

\leavevmode\hypertarget{ref-MahachieJohn2011}{}%
29. Mahachie John JM, Van Lishout F, Van Steen K. Model-Based Multifactor Dimensionality Reduction to detect epistasis for quantitative traits in the presence of error-free and noisy data. Eur J Hum Genet. 2011; doi:\href{https://doi.org/10.1038/ejhg.2011.17}{10.1038/ejhg.2011.17}.

\leavevmode\hypertarget{ref-VanLishout2013}{}%
30. Lishout FV, Mahachie John JM, Gusareva ES, Urrea V, Cleynen I, Théâtre E, et al. An efficient algorithm to perform multiple testing in epistasis screening. BMC Bioinformatics. 2013; doi:\href{https://doi.org/10.1186/1471-2105-14-138}{10.1186/1471-2105-14-138}.

\leavevmode\hypertarget{ref-MahachieJohn2012}{}%
31. Mahachie John JM, Cattaert T, Van Lishout F, Gusareva ES, Van Steen K. Lower-Order Effects Adjustment in Quantitative Traits Model-Based Multifactor Dimensionality Reduction. PLoS One. 2012; doi:\href{https://doi.org/10.1371/journal.pone.0029594}{10.1371/journal.pone.0029594}.

\leavevmode\hypertarget{ref-Lishout2015}{}%
32. Lishout FV, Gadaleta F, Moore JH, Wehenkel L, Steen KV, Lishout V, et al. gammaMAXT: a fast multiple-testing correction algorithm. BioData Min. 2015; doi:\href{https://doi.org/10.1186/s13040-015-0069-x}{10.1186/s13040-015-0069-x}.

\leavevmode\hypertarget{ref-Gola2018}{}%
33. Gola D. MBMDRClassifieR. 2018. \url{https://github.com/imbs-hl/MBMDRClassifieR}. Accessed 1 Nov 2018.

\leavevmode\hypertarget{ref-Zou2005}{}%
34. Zou H, Hastie T. Regularization and variable selection via the elastic net. J R Stat Soc B. 2005; doi:\href{https://doi.org/10.1111/j.1467-9868.2005.00503.x}{10.1111/j.1467-9868.2005.00503.x}.

\leavevmode\hypertarget{ref-RCoreTeam2015}{}%
35. R Core Team. R: A Language and Environment for Statistical Computing. 2016. \url{http://www.r-project.org/}. Accessed 1 Nov 2018.

\leavevmode\hypertarget{ref-Wright2017}{}%
36. Wright MN, Ziegler A. ranger: A Fast Implementation of Random Forests for High Dimensional Data in C++ and R. J Stat Softw. 2017; doi:\href{https://doi.org/10.18637/jss.v077.i01}{10.18637/jss.v077.i01}.

\leavevmode\hypertarget{ref-Bischl2016}{}%
37. Bischl B, Lang M, Richter J, Bossek J, Judt L, Kuehn T, et al. mlr: Machine Learning in R. J Mach Learn Res. 2016;17:1--5.

\leavevmode\hypertarget{ref-Bischl2017}{}%
38. Bischl B, Richter J, Bossek J, Horn D, Thomas J, Lang M. mlrMBO: A Modular Framework for Model-Based Optimization of Expensive Black-Box Functions. 2017. \url{http://arxiv.org/abs/1703.03373}.

\leavevmode\hypertarget{ref-Amos2009}{}%
39. Amos CI, Chen W, Seldin MF, Remmers EF, Taylor KE, Criswell LA, et al. Data for Genetic Analysis Workshop 16 Problem 1, association analysis of rheumatoid arthritis data. BMC Proc. 2009; doi:\href{https://doi.org/10.1186/1753-6561-3-s7-s2}{10.1186/1753-6561-3-s7-s2}.

\leavevmode\hypertarget{ref-Kruppa2012}{}%
40. Kruppa J, Ziegler A, König IR. Risk estimation and risk prediction using machine-learning methods. Hum Genet. 2012; doi:\href{https://doi.org/10.1007/s00439-012-1194-y}{10.1007/s00439-012-1194-y}.

\leavevmode\hypertarget{ref-Liu2011}{}%
41. Liu C, Ackerman HH, Carulli JP. A genome-wide screen of gene--gene interactions for rheumatoid arthritis susceptibility. Hum Genet. 2011; doi:\href{https://doi.org/10.1007/s00439-010-0943-z}{10.1007/s00439-010-0943-z}.

\end{document}